\documentclass[journal]{IEEEtran}

\usepackage{times}
\usepackage{mathrsfs}
\usepackage{epsfig}
\usepackage{graphicx}
\usepackage{amsmath}
\usepackage{amssymb}
\usepackage{graphicx}
\usepackage{cite}
\usepackage{enumerate}
\usepackage{cases}
\usepackage{multirow}
\usepackage{verbatim}
\usepackage{amssymb}
\usepackage{CJK}
\usepackage{algorithm}
\usepackage{algorithmicx}
\usepackage{algpseudocode}
\usepackage{color}
\usepackage{bm}
\usepackage{booktabs}
\usepackage{amssymb}
\usepackage{makecell}
\usepackage{url}

\usepackage{subfig}
\usepackage{color}
\usepackage{tikz}
\usepackage{ifthen}
\usepackage{bm}
\usetikzlibrary{3d,decorations.text,shapes.arrows,positioning,fit,backgrounds, intersections}
\usetikzlibrary{shapes.misc}

\usepackage{multirow}
\usepackage{enumitem}

\usepackage{booktabs}

\newcommand{\etal}{\textit{et al}.}
\newcommand{\ie}{\textit{i}.\textit{e}.}
\newcommand{\eg}{\textit{e}.\textit{g}.}
\newcommand{\etc}{\textit{etc}}

\begin{document}

\title{Learning Hierarchical Color Guidance for Depth Map Super-Resolution}
\author
{
Runmin Cong,~\IEEEmembership{Senior Member,~IEEE,} Ronghui Sheng, Hao Wu, Yulan Guo, Yunchao Wei, Wangmeng Zuo, \\ Yao Zhao,~\IEEEmembership{Fellow,~IEEE,} and Sam Kwong,~\IEEEmembership{Fellow,~IEEE}

\thanks{Runmin Cong is with the Institute of Information Science, Beijing Jiaotong University, Beijing 100044, China, also with the School of Control Science and Engineering, Shandong University, Jinan 250061, China, and also with the Key Laboratory of Machine Intelligence and System Control, Ministry of Education, Jinan 250061, Shandong, China (e-mail: rmcong@sdu.edu.cn).}
\thanks{Ronghui Sheng, Yunchao Wei, and Yao Zhao are with the Institute of Information Science, Beijing Jiaotong University, Beijing 100044, China (e-mail: ronghuisheng@bjtu.edu.cn; wychao1987@gmail.com; yzhao@bjtu.edu.cn).}
\thanks{Hao wu is with the Artificial Intelligence Institute, Beijing Normal University, Beijing 100875, China (email: wuhao@bnu.edu.cn).}
\thanks{Yulan Guo is with the College of Electronic Science and Technology, National University of Defense Technology (NUDT), Changsha 410073, China, and also with the School of Electronics and Communication Engineering, Sun Yat-sen University, Guangzhou 510275, China (e-mail:guoyulan@sysu.edu.cn).}
\thanks{Wangmeng Zuo is with School of Computer Science and Technology, Harbin Institute of Technology, Harbin, 150001, China (e-mail: wmzuo@hit.edu.cn).}
\thanks{Sam Kwong is with the Lingnan University, Hong Kong SAR, China (e-mail: samkwong@ln.edu.hk).}
}

\markboth{IEEE TRANSACTIONS ON INSTRUMENTATION AND MEASUREMENT}
{Shell \MakeLowercase{\textit{et al.}}: Bare Demo of IEEEtran.cls for IEEE Journals}
\maketitle

\begin{abstract}
Color information is the most commonly used prior knowledge for depth map super-resolution (DSR), which can provide high-frequency boundary guidance for detail restoration. However, its role and functionality in DSR have not been fully developed. In this paper, we rethink the utilization of color information and propose a hierarchical color guidance network to achieve DSR. On the one hand, the low-level detail embedding module is designed to supplement high-frequency color information of depth features in a residual mask manner at the low-level stages. On the other hand, the high-level abstract guidance module is proposed to maintain semantic consistency in the reconstruction process by using a semantic mask that encodes the global guidance information. The color information of these two dimensions plays a role in the front and back ends of the attention-based feature projection (AFP) module in a more comprehensive form. Simultaneously, the AFP module integrates the multi-scale content enhancement block and adaptive attention projection block to make full use of multi-scale information and adaptively project critical restoration information in an attention manner for DSR. Compared with the state-of-the-art methods on four benchmark datasets, our method achieves more competitive performance both qualitatively and quantitatively. The code and results can be found from the link of \url{https://rmcong.github.io/HCGNet\_TIM2024}.
\end{abstract}


\IEEEpeerreviewmaketitle

\section{Introduction} \label{sec1}
\IEEEPARstart{D}{EPTH} maps describe the distance relationship of the scene including the occlusion and overlap of objects, which is essential for the 3D understanding tasks, such as autonomous driving \cite{TIM-drive}, 3D reconstruction \cite{TIM-3D,TIM-3D2}, object recognition \cite{TIM-recognition,TIM-recognition2}, and salient object detection \cite{SOD-23-CSVT,crm/tip22/CIRNet,crm/tip21/DynamicRGBDSOD,crm/acmmm21/CDINet,crm/ACMMM23/PICRNet,crm/tmm22/3DSaliency,SOD-22-TGRS,crm/tmm22/TNet,crm/tcyb21/ASIFNet,crm/eccv20/RGBDSOD}, \etc. 
However, due to the limitations of existing depth acquisition devices, the resolution of the acquired depth maps is relatively low, especially for the low-power depth sensors equipped on smartphones. The low-resolution (LR) depth map cannot match the high-resolution (HR) color image in resolution, thereby hindering the further expansion of depth-oriented applications. Therefore, the super-resolution reconstruction technology for depth maps came into being, which has practical research value and industrial application value.\\ 
\begin{figure}[t]
	\centering
	\includegraphics[width=0.48\textwidth]{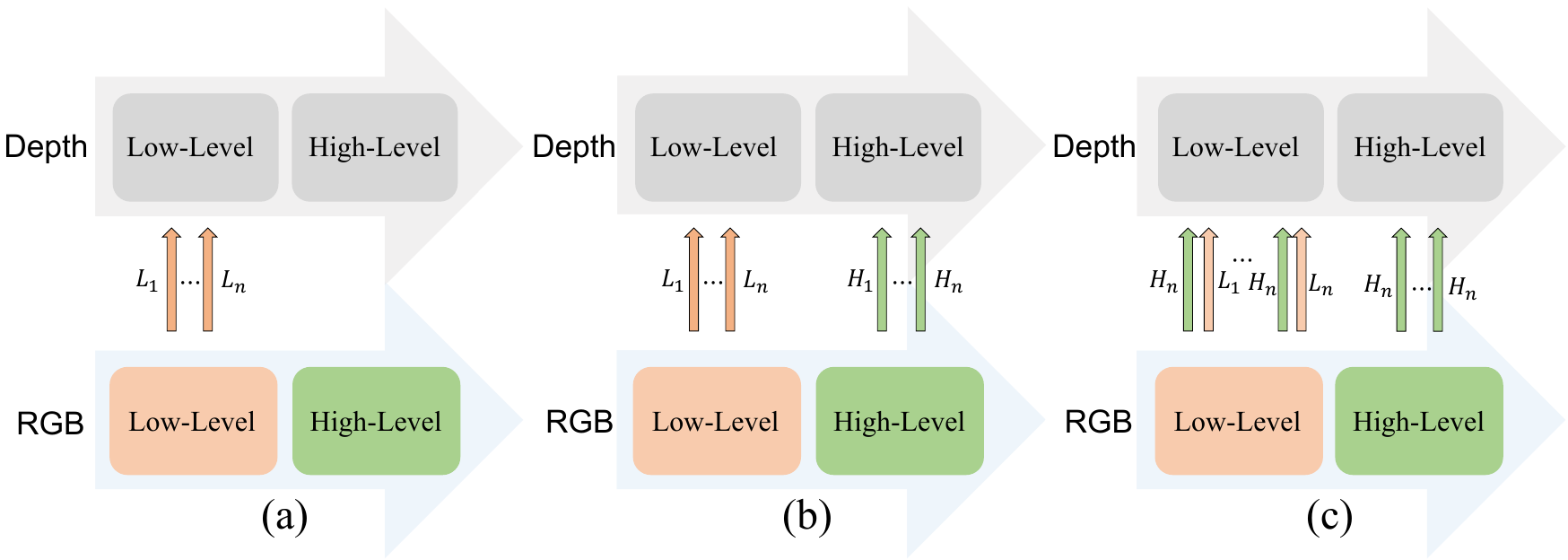}		
    \caption{Illustration of the color guidance in DSR. Mode (a) only utilizes the low-level color information to guide the reconstruction of detail information; Mode (b) treats different levels of color information indiscriminately; Mode (c) represents our guidance model, which divides color information into two parts, \ie, low-level and high-level information, and allows them to play different roles. }
    \vspace{-0.25cm}
	\label{fig1}
\end{figure}
\ \ Depth map super-resolution (DSR) is a challenging task that aims to reconstruct the LR depth map into an HR depth map. This task is inherently an ill-posed inverse problem due to the absence of a unique mapping between LR and HR depth maps. Furthermore, it is particularly difficult to recover fine details, such as sharp boundaries, especially when dealing with large upsampling factors \cite{cvpr2020,WAFP,crm/access17/dsr}. 
Because of the structural similarity with depth maps and readily accessible, HR color images can naturally provide comprehensive guidance information for the DSR task, and numerous color-guided DSR approaches have been proposed. 
However, what kind of color information is to be utilized for guidance and how the implementation is to be conducted still remain open topics in color-guided DSR. 
For example, current DSR techniques utilize the color boundary information, either explicitly or implicitly, to enhance the reconstruction of details \cite{DKN,PDR,zuo2020,BridgeNet}. 
But such structural congruence is not universally applicable. The RGB image contains not only the object boundary but also the texture boundary inside the object, while the depth map only has the object boundary. 
In other words, as for the color-guided DSR, a critical issue is the effective exploitation of color guidance information to enhance depth details while mitigating the texture-copying artifacts introduced by the color image.
Furthermore, in the form of guidance, some modes and strategies are designed, such as using low-level color features as detailed guidance, usually directly concatenating color features with depth features \cite{Simultaneous-RGBD-SR,PMBANet} (as shown in Fig.  \ref{fig1}(a)), or treating different levels of color features equally as guidance \cite{guo2019,DAEA} (as shown in Fig.  \ref{fig1}(b)), \etc. However, these methods do not fully consider the roles and diversity of different color information in the guidance phase, indicating a necessity for more in-depth and comprehensive exploration to leverage the full spectrum of color guidance information effectively.


Motivated by the above analysis, 
the core theoretical contribution of our work lies in rethinking the utilization of color information in the DSR task and distinguishing the roles of low-level and high-level color information, thereby making them guide the depth branch in a divide-and-conquer manner, as shown in Fig. \ref{fig1}(c).
On the one hand, the low-level color features contain fine-grained detailed information (\eg, boundaries) \cite{2020SOD, PMBANet} that DSR needs to pay attention to, which is helpful for the detail recovery of the depth map. 
However, the representation of these features is too specific, with a lot of interference. And simply transferring color features may introduce unnecessary interfering boundaries, resulting in texture replication. 
To address this, we learn a residual mask in the designed low-level detail embedding (LDE) module to highlight the spatial locations of color features that are most consistent with depth features, thereby adaptively guiding the information transmission from color features to depth features.
On the other hand, the high-level color features contain global abstract information, which describes scene content more comprehensively and preserves semantic outlines. 
The existing approaches do not specifically consider high-level color abstract information, but discard it \cite{wen2019} or treat it the same as low-level detail information \cite{guo2019}. Considering that the semantic consistency of the scene may be shifted or blurred during the depth reconstruction, we design a high-level abstract guidance (HAG) module to modify the initial reconstruction features by using a semantic mask that encodes the global abstract guidance information. It is worth mentioning that the LDE and HAG modules we designed also have good portability, which can be transplanted into existing color-guided DSR methods to improve their performance (see the validation experiments in Section \ref{ab}). 

In addition, to achieve better recovery, we need to map the low-resolution features to the desired high-resolution reconstruction features. The existing methods, such as DBPN \cite{guo2019}, directly map features between LR and HR domains instead of selecting the reconstruction region, which greatly increases the complexity of the model and introduces additional errors. In fact, the focus of the DSR task is not to generate content from scratch but to supplement, refine, and enhance the details such as boundaries. From this point of view, blindly and indiscriminately performing super-resolution reconstruction on all regions is a sub-optimal way, which also difficult to achieve the purpose of optimizing important regions with more severe degradation. To this end, we design the attention-based feature projection (AFP) module, including a multi-scale content enhancement (MCE) block and multiple adaptive attention projection (AAP) blocks. The core contribution of the AFP module lies in the designed AAP block, which reinforces the important restoration regions in an attention manner, thereby suppressing interference and improving the reconstruction performance.
The whole reconstruction process is a restoration pipeline from coarse to fine, focusing on using different levels of color information for guided reconstruction. All modules cooperate with each other to hierarchically reconstruct the depth features, thereby obtaining the final depth map at the target resolution.

To summarize, the contributions of this work are as follows:
\begin{itemize}[noitemsep, topsep=0pt]
	\item We reexplore the role of the color information in DSR and propose a hierarchical color guidance network (HCGNet). Comprehensive experiments on four benchmark datasets show that the proposed method achieves more competitive performance both qualitatively and quantitatively.
    \item The LDE and HAG modules work together to achieve hierarchical color guidance in DSR task. Concretely, the LDE module distinguishes between similar as well as interfering regions in the form of residual masks, thereby effectively utilizing high-frequency complementary guidance of color features. And the HAG module extracts complete semantic outlines from high-level abstract features in the form of semantic masks, thereby alleviating semantic shifts and ambiguities for global reconstruction.
	\item 
	We design an AAP block to reinforce the key restoration regions in the attention domain, thereby suppressing the valueless redundancy and improving the reconstruction performance with optimized computation.
\end{itemize}
\section{Related Work}

\subsection{Non Color-Guided DSR}
Non color-guided DSR directly reconstructs a HR depth map from a LR depth map without any external guidance information. Earlier works proposed some local filtering-based methods, which mostly use high-pass filters to recover the boundary information of the depth map. For example, Yang \etal \cite{Yang2007} proposed a post-processing method to improve the spatial resolution and accuracy of depth images by iterative bilateral filtering. The filtering-based method has lower operational complexity, but its ability to recover depth details is unsatisfactory, and the over-smooth and blurred boundaries are prone to appear in the reconstructed depth map. In recent years, with the development of deep learning \cite{DBPN, TIM-DEWRN, TIM-SRN, zhang2023controlvideo}, the research focus has gradually shifted to deep learning solutions for SR, and many high-performance algorithms have emerged, such as DBPN \cite{DBPN}, DEWRN \cite{TIM-DEWRN}, SRN \cite{TIM-SRN}, SADN \cite{TIM-SADN}, DTSR \cite{TIM-DTSR}, \cite{mvp1, mvp2}, \etc.\ Taking into account the particularity of the depth map, deep learning-based DSR methods usually needs to design a specific network structure to improve the reconstruction performance. Riegler \etal \cite{ATGVNet} incorporated the total generalized variational constraint at the back-end of DCNN to form an end-to-end ATGV-Net. Song \etal \cite{song2019} proposed to reconstruct the depth map in a way that a series of view synthesis sub-tasks can be learned in parallel. Sun \etal \cite{sun2020acm} proposed a depth-controlled slicing network that learns a set of slicing branches in a divide-and-conquer manner and is parameterized by a distance-aware weighting scheme to adaptively integrate the different depths in the set.
\begin{figure*}[t]
	\centering
	\includegraphics[width=0.98\textwidth,
 height=0.54\textwidth]{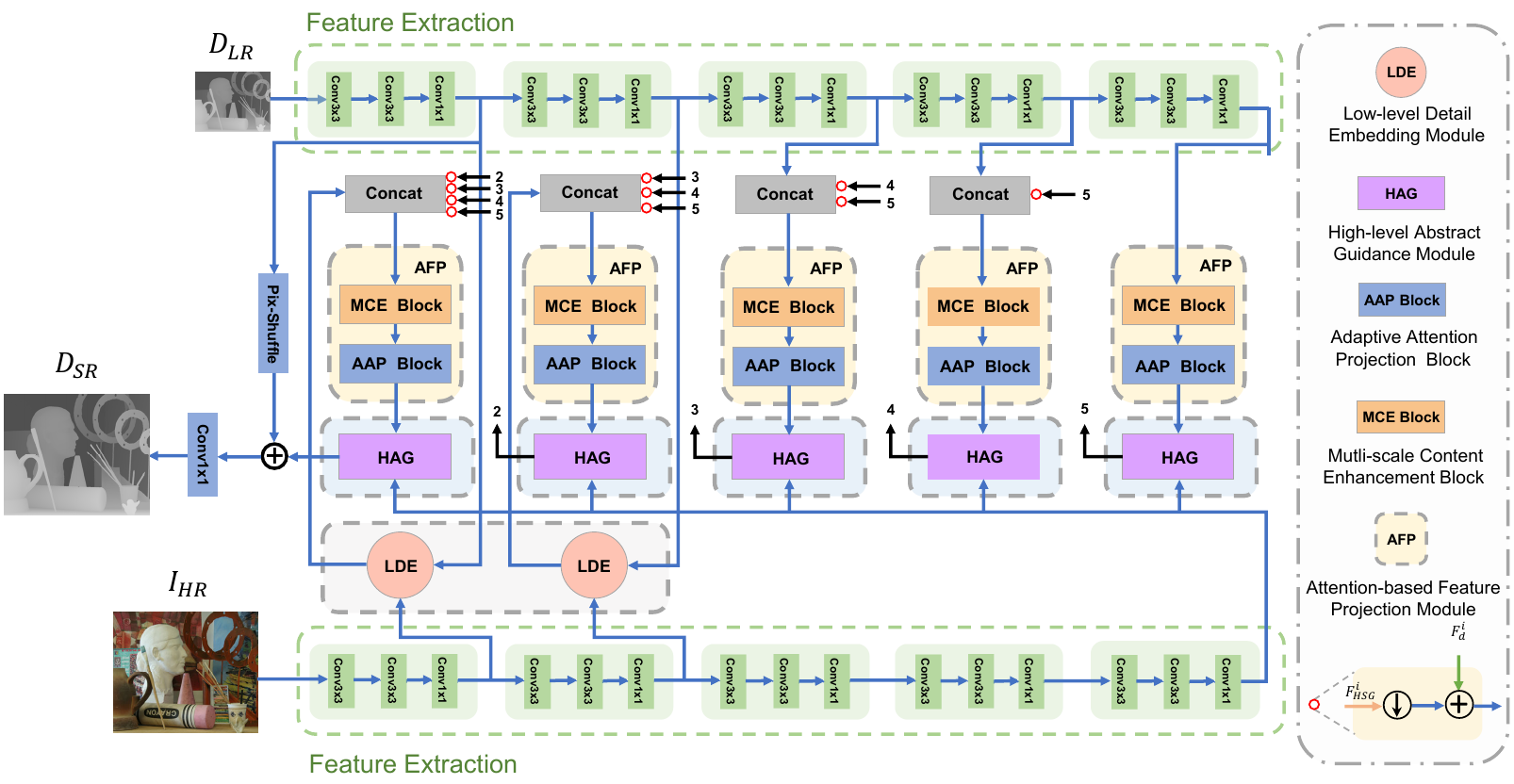}	
	\caption{The architecture of HCGNet. The LR depth map and HR color image are first embedded into the feature extraction unit to extract multi-level features. Then, the Attention-based Feature Projection (AFP) module, Low-level Detail Embedding (LDE) module, and High-level Abstract Guidance (HAG) module work together to gradually recover details in LR depth features and generate the HR depth map. The use of color information is manifested in two aspects. On the one hand, the low-level color features are used in the low-level reconstruction stage to restore details through the LDE module. On the other hand, the high-level abstract features are used at the end of the AFP module to provide semantic guidance through the HAG module. }
	
	\label{fig: main}
 \vspace{-0.15cm}
\end{figure*}
\subsection{Color-Guided DSR}
As mentioned earlier, it is effortless for some depth cameras (such as Kinect) to acquire HR color images while acquiring depth maps. Therefore, the color-guided DSR model has received widespread attention in recent years and has become the mainstream model. The color-guided DSR is based on the similar structural information between the depth map and the aligned color image, \ie, the depth boundaries have a strong symbiotic relationship with the luminance boundaries. Filter-based approaches consider coeval structural relationships in addition to depth-neighborhood relationships when designing filters. For example, Kopf \etal \cite{kopf} proposed a joint bilateral upsampling filter model by combining a Gaussian function based on the depth image neighborhood position relationship. He \etal \cite{He2013} developed a local linear model of the filtered image and the bootstrap image model, and then proposed a bootstrap filter.
Wang \etal \cite{DNR} proposed a dual normal-depth regularization term to constrain the edge consistency between the normal map and the depth map. Recently, learning-based approaches have successfully applied DCNN to the field of color-guided DSR. Wen \etal \cite{wen2019} used a coarse-to-fine DCNN network to learn different filters with different kernel sizes, thus enabling data-driven training to replace the manually designed filters. Huang \etal \cite{huang2019} proposed a deep dense residual network with a pyramidal structure that leverages multi-scale features to predict high-frequency residuals through dense connectivity and residual learning. Guo \etal \cite{guo2019} designed a residual U-Net structure for the deep reconstruction task and introduced hierarchical feature-driven residual learning. Zuo \etal \cite{zuo2020} proposed a data-driven super-resolution network based on global and local residual learning. Sun \etal \cite{PMBANet} proposed a progressive multi-branch aggregation network that utilizes multi-scale information and high-frequency features to fully reconstruct the depth map in a progressive way. They also demonstrate that the low-level color information is only suitable for early feature fusion and does not help much for DSR at $\times$2 and $\times$4 cases.

\section{Methodology}
\subsection{Overview}
The overview of the proposed network is shown in Fig. \ref{fig: main}, which is a dual-stream hierarchical reconstruction architecture. Given a LR depth map $D_{LR} \in \mathbb{R}^{h\times w\times 1}$ and the corresponding HR color image $I_{HR} \in \mathbb{R}^{H\times W\times 3}$ as inputs, the goal of our task is to reconstruct and generate a SR version of depth map $D_{SR} \in \mathbb{R}^{H\times W\times 1}$ with the same resolution of the color image. 
To be concise, we first extract the multi-level features of RGB and depth features via five progressive convolution blocks (green blocks in Fig. \ref{fig: main}), where each block includes two $3\times 3$ convolution layers and a $1\times 1$ convolution layer. 
The obtained RGB and depth features are denoted as $\mathrm{F}_c^i$ and $\mathrm{F}_d^i$ ($i=\{1,2,3,4,5\}$), respectively.

Then, we achieve color-guided depth feature learning and detail restoration under the cooperation of the AFP, LDE, and HAG modules. It is worth noting that there are three inputs (if any) are sent to the AFP module: 
(1) The depth backbone features $\mathrm{F}_d^i$ in the corresponding level; 
(2) The low-level detail features $\mathrm{F}_{LDE}^i$ generated by the LDE module, which is used for detailed restoration in the low-level reconstruction stage; 
(3) The dense transfer features ($\mathrm{F}_{tr}^{i+1},\mathrm{F}_{tr}^{i+2},\cdots,\mathrm{F}_{tr}^5$) from all the completed reconstruction levels. 
At different reconstruction levels, the input features of the AFP module are different, which are specifically formulated as:
\begin{equation}
	\mathrm{F}_{in}^i = \left\lbrace 	 \
	\begin{aligned}
	    &\mathrm{F}_d^i, &&i=5 \\
		Concat&(\mathrm{F}_d^i, \mathrm{F}_{tr}^k), &&i= \{3,4\} \\
		Concat&(\mathrm{F}_{LDE}^i, \mathrm{F}_{tr}^k), &&i= \{1,2\} \\
	\end{aligned}\right. ,
\end{equation}
where $Concat$ denotes the concatenation operation along the channel dimension, $\mathrm{F}_d^i$ represent the depth backbone features of the $i$-th level, $\mathrm{F}_{LDE}^i$ are the low-level detail features generated by the $i$-th LDE module, and $\mathrm{F}_{tr}^k$ denote the transfer features from the $k$-th completed reconstruction level, which can be calculated by:
\begin{equation}
    \mathrm{F}_{tr}^k = (\mathrm{F}_{HAG}^k)\downarrow + \mathrm{F}_d^k,
    \label{eq2}
\end{equation}
where $\mathrm{F}_{HAG}^k$ represent the output features of the $k$-th HAG module, $\downarrow$ denotes the downsampling operation, and  $k=\{i+1,i+2,\cdots,5\}$.
It should be noted that the inputs of the LDE module include the depth features $\mathrm{F}_d^i$ and color features $\mathrm{F}_c^i$ of the corresponding layer.

After that, the high-level abstract features $\mathrm{F}_c^5$ and the depth backbone features $\mathrm{F}_d^i$ are fed into the HAG module to modify the output features $\mathrm{F}_{AFP}^{i}$ of AFP module and generate the reconstruction features $\mathrm{F}_{HAG}^{i}$. 
Finally, the pixel-shuffle and convolution operations are performed on the features $\mathrm{F}_{d}^{1}$ and $\mathrm{F}_{HAG}^{1}$ to obtain the final upsampled depth map $D_{SR}$.

Note that, our model is trained by minimizing $L_1$ loss, which can be formulated as:
\begin{equation}
    Loss = \left\|D_{SR}-D_{HR}\right\|_1 ,
\end{equation}
where $D_{SR}$ and $D_{HR}$ denote the predicted depth SR result and ground truth, respectively, and $\left\|\cdot\right\|_1$ is the $L_1$ norm function. In the following subsections, we will provide the technical details of the AFP, LDE, and HAG modules one by one.

\subsection{Attention-based Feature Projection Module}

In order to achieve the depth map super-resolution, we need to map the low-resolution features to the desired high-resolution reconstruction features. Specifically, there are two issues that need to be paid attention to: (1) In order to recover more severely degenerated local details (such as depth boundaries and fine objects), simply increasing the depth of the network is insufficient and unwise. Therefore, we introduce a Multi-scale Content Enhancement (MCE) block to enhance the depth features before projection, using different receptive fields to recover detailed features at different scales as much as possible.
(2) The information between the LR and HR domains is not absolutely one-to-one correspondence in the projection process, and the interference of excessive interfering information is likely to introduce additional errors, thereby impairing the reconstruction accuracy. To this end, we propose an Adaptive Attention Projection (AAP) block to project valid information in the attention domain, guaranteeing the effectiveness and compactness of the projected features. Note that, four cascaded AAP blocks are used in the AFP module to achieve better performance. In summary, as shown in Fig. \ref{fig:3}, the MCE block and AAP blocks together form the AFP module to achieve depth feature reconstruction.
	

\subsubsection{Multi-scale Content Enhancement block} 
\begin{figure*}[!ht]
	\centering
	\includegraphics[width=1\textwidth]{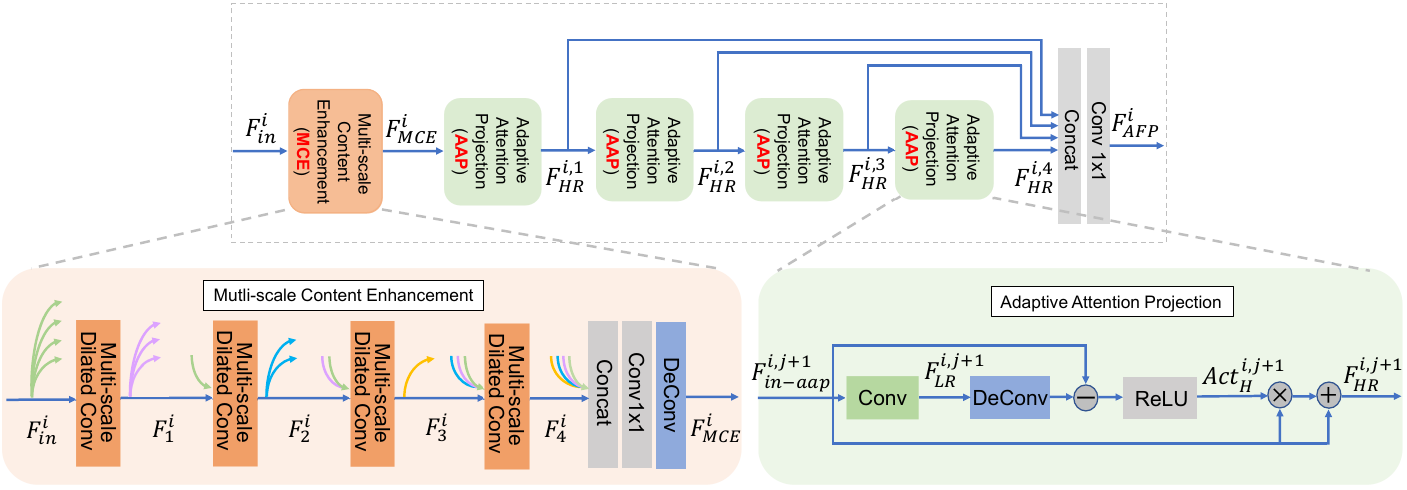}		
	\caption{The whole architecture of AFP module and details of sub-blocks, \ie, MCE block and AAP block.}
	
	\label{fig:3}
\end{figure*}

Multi-scale information can effectively perceive and model different details, which is of great significance for detail restoration in DSR. As shown in the lower left of Fig. \ref{fig:3}, the MCE block contains a stack of four dilated convolution layers with different dilation rates, which are applied to capture more details with different scales of receptive fields \cite{dilated,crm/tip20/MCMT-GAN}. Moreover, we employ dense connections to obtain full information from all previous layers, which are formulated as:
\begin{align}
	\mathrm{F}_m^i = \left\lbrace 	 \
	\begin{aligned}
	    &MD(\mathrm{F}_{in}^i),  &&m = 1 \\
		&MD(Concat(\mathrm{F}_{in}^i,\mathrm{F}_1^i,\cdots,\mathrm{F}_{m-1}^i)), &&m= \{2,3,4\} \\
	\end{aligned}\right.  
\end{align}
where $MD$ denotes the multi-scale dilated convolution operation with different dilation rates of 1, 2, 3, and 4, and $\mathrm{F}_m^i$ is the output of each multi-scale dilated convolution. Finally, all the multi-scale dilation features are concatenated and fused by a $1\times 1$ convolution layer:
\begin{align}
	\mathrm{F}_{MCE}^i = DeConv(Conv_{1 \times 1}(Concat(\mathrm{F}_{in}^i,\mathrm{F}_1^i,\cdots,\mathrm{F}_4^i))), 
\end{align} 
where $Conv_{1 \times 1}$ denotes the convolution layer with the kernel size of $1\times 1$, $\mathrm{F}_{MCE}^i$ is the output of MCE block, which perceives content information of different scales, and $DeConv$ denotes the upsampling operation implemented by the deconvolution layer. 
\subsubsection{Adaptive Attention Projection block} 
The super-resolution process of depth maps needs to bridge the huge gap between the LR domain and the HR domain. 
In fact, the focus of the DSR task is not to generate content from scratch but to supplement, refine, and enhance the details such as boundaries. 
From the perspective of the frequency domain, low-frequency information is usually included in the smooth regions while high-frequency regions contain more boundary information. Therefore, to extract clear color boundaries and suppress their interfering textures, we need to correct the error information progressively while extracting the high-frequency features of the image. 
Moreover, blindly and indiscriminately performing super-resolution reconstruction on all regions is a sub-optimal way, which is also difficult to achieve the purpose of optimizing important regions with more severe degradation.
In other words, in the process of restoring information from the LR domain to the HR domain (which is also called the projection process), interference may be introduced without filtering, thereby introducing additional errors and affecting the reconstruction accuracy.
Hence, we design the AAP block to reinforce the key restoration regions in an attention manner, thereby suppressing the interference and improving the reconstruction performance, as shown in the bottom flowchart of Fig. \ref{fig:3}.

 

For the AAP blocks, the input of the first AAP block is the output features of the MCE block, while the input of the other blocks is the output of the previous AAP block. As such, the input of the AAP block can be uniformly formulated as:
\begin{align}
	\mathrm{F}_{in-aap}^{i,j+1} &= \left\lbrace 	 \
	\begin{aligned}
	    &\mathrm{F}_{MCE}^i,   
        &&j = 0 \\
		&\mathrm{F}_{HR}^{i,j},  &&j= \{1,2,3\} \\
	\end{aligned}\right.
	\label{eq6}
\end{align} 
where $\mathrm{F}_{HR}^{i,j}$ is the HR output of $j$-th AAP block in the $i$-th level (will be formulated further below). 

For the algorithm of AAP, we simulate the DSR process by using both down-projection and up-projection, thereby obtaining the reconstructed HR feature map under worse conditions. These projection blocks can be interpreted as self-correcting processes that provide projection errors to the sampling layer, and thus progressively generate better solutions:

\begin{equation}
	\mathrm{F}_{{HR}_{rough}}^{i,j+1} = DeConv(Conv(\mathrm{F}_{in-aap}^{i,j+1})),
\end{equation}
where $Conv$ is a convlution layer for down-projection, and $DeConv$ is a deconvolution layer for up-projection. 

Then, we subtract the reconstructed HR features from the original HR features to generate the residual features and extract the high-frequency features of the image, which encode the content information that needs to be recovered during reconstruction. The projected attention map is calculated by:
\begin{equation}
	\mathrm{Act}_H^{i,j+1} = ReLU(\mathrm{F}_{in-aap}^{i,j+1} - \mathrm{F}_{{HR}_{rough}}^{i,j+1}),
\end{equation}
where 
$ReLU$ denotes the rectified linear unit. The projected attention map will correct errors in reconstruction and avoid the degradation caused by feature projection between the LR and HR domains.

Finally, the residual map is activated as a projected attention map and used to adaptively refine the original HR features:
\begin{equation}
	\mathrm{F}_{HR}^{i,j+1} = \mathrm{Act}_H^{i,j+1} \otimes \mathrm{F}_{in-aap}^{i,j+1} + \mathrm{F}_{in-aap}^{i,j+1},
\end{equation}
where $\otimes$ denotes the element-wise multiplication. 
With four serial AAP blocks, four HR reconstruction features from coarse to fine are generated. Combining them, we can obtain the final output of the AFP module:
\begin{equation}
	\mathrm{F}_{AFP}^{i} = Conv_{1 \times 1}(Concat(\mathrm{F}_{HR}^{i,1},\mathrm{F}_{HR}^{i,2},\mathrm{F}_{HR}^{i,3},\mathrm{F}_{HR}^{i,4})),
\end{equation}
where $\mathrm{F}_{AFP}^{i}$ denote the initial reconstruction depth features.

\subsection{Low-level Detail Embedding Module}

As is well-known, high-resolution color images are readily available and contain much useful information, such as boundaries, textures, and semantic information, \etc. 
Therefore, introducing color guidance into the DSR model has become the mainstream idea in this field.
However, there is no complete consensus on which color information to use and how to use it. Considering the different roles of color features at different levels, we provide a differentiated solution of color guidance strategy in this paper. Concretely, we design a Low-level Detail Embedding (LDE) module at the low-level reconstruction stage to leverage low-level color features for enhancing the high-frequency details of depth features, such as boundaries. In addition, we design a High-level Abstract Guidance (HAG) module, where the high-level abstract color features are used to perform content correction on the original reconstruction features, preventing content shifts during the depth reconstruction. We will introduce the LDE module in this subsection, and provide the details of the HAG module in the next subsection.

\begin{figure}[!t]
    \setlength{\abovecaptionskip}{8pt}
	\centering
	\includegraphics[width=0.48\textwidth]{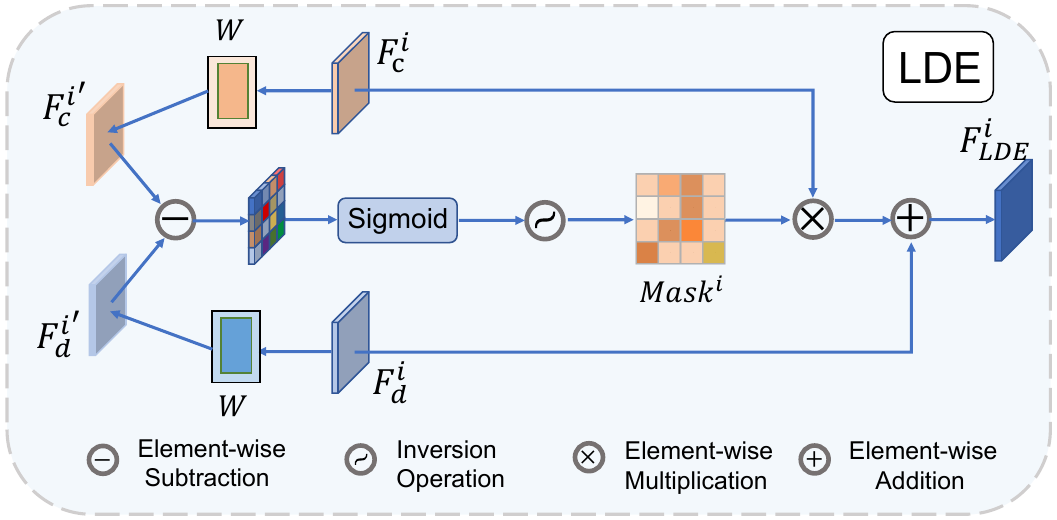}
	\caption{The architecture of Low-level Detail Embedding Module.}
	\label{fig5.1}
	\vspace{-0.15cm}
\end{figure}

For depth map super-resolution, accurate and sharp boundary reconstruction has always been the focus of researchers' unremitting efforts. It just so happens that what the low-level layer of the color branch learns is detailed information such as texture, boundary, \etc. 
Therefore, we introduce the color features at lower levels (\ie, the first two layers) of the HR color branch through the LDE module and take the output as one of the inputs of the AFP module. 
However, the depth boundaries are not absolutely consistent with the RGB boundaries. In fact, the boundaries in the depth map are mainly the object boundaries, while the color image includes rich texture boundaries inside the object in addition to the object boundaries. Obviously, the texture boundaries are interferences for DSR. However, it is difficult to determine the delineation of texture and boundaries very clearly in network reconstruction, so completely discarding texture details is not the best option. 
Therefore, instead of removing all texture information with absolute gradient boundaries, our proposed LDE module suppresses the interfering information of RGB by learning residual masks, which is shown in Fig. \ref{fig5.1}. 


Concretely, we first make a domain mapping of the RGB-D features using point-wise convolution and depth-wise convolution. The subsequent key step of subtraction actually serves to distinguish between similar as well as interfering regions in the RGB-D features. Since explicitly filtering out the appropriate complementary regions is difficult, we obtain the regions with large variances in the RGB-D modality, \ie, the redundant regions, by subtractive operations. The reverse residual mask, unlike the binarized mask, will increase the weight of complementary information and reduce the weight of information in redundant regions, thus differentiating the guiding role of low-level color features:
\begin{equation}
	\mathrm{RM}^i = 1-Sigmoid(Conv_{1 \times 1}(W_c\cdot \mathrm{F}_c^i - W_d\cdot \mathrm{F}_d^i)),
\end{equation}
where $\mathrm{RM}^i$ denotes the residual mask, $W_c$ and $W_d$ represent the mapping matrix for the color features and depth features, 
$Sigmoid$ is the normalization operation, and $i$ indexes the lower level here, which is equal to 1 or 2.

In this way, the residual mask highlights the most relevant part of the color and depth information, so we multiply it with the initial color features to obtain the effective color features that can be used for depth reconstruction guidance. 
Moreover, we believe that the feature representation of the color information filtered by the residual mask in the lower levels is in the same domain as the depth information, so the final fusion adopts a direct summation scheme, which is also consistent with objective rules.
Therefore, the final output of the LDE module can be formulated as:
\begin{equation}
\mathrm{F}_{LDE}^i = \mathrm{RM}^i \otimes \mathrm{F}_c^i + \mathrm{F}_d^i,
\end{equation}
where $\otimes$ denotes the element-wise multiplication. 

\subsection{High-level Abstract Guidance Module}
As analyzed earlier, the existing methods mainly focus on extracting color features to supplement the details for depth reconstruction, just like the functions implemented by our LDE module. However, let us rethink the role of color-guided features: Is this detailed guidance strategy sufficient? In fact, high-level color features are very important for many tasks, 
which contains abstract global information and preserves semantic outlines.
In the DSR task, the existing methods ignore one issue, \ie, the global abstract information preserving ability of the reconstructed features. As the reconstruction process progresses, there is a possibility that semantic consistency may be shifted or blurred, which is very unfavorable for the subsequent depth-oriented application tasks. This is mainly caused by the lack of global guidance in the reconstruction. 
Fittingly, our color branch can provide high-resolution, offset-free color guidance information. This also benefits from the fact that the semantic information extraction for the color branch does not rely on the process of depth reconstruction, but extracts shallow texture features and high-level abstract features through multi-layer convolution deepening. Since no resolution change is involved, the high-resolution color information is not shifted during the extraction process.
Inspired by these, we design a HAG module to maintain the content attribute during the depth reconstruction, which is equipped after each AFP module. Specifically, the high-level color information at the top layer of the color branch is utilized to generate a mask that encodes the global content guidance information, and is further used to modify the initial reconstruction features $\mathrm{F}_{AFP}^{i}$ (\ie, the output of the AFP module).

\begin{figure}[t]
    \setlength{\abovecaptionskip}{8pt}
	\centering
	\includegraphics[width=0.48\textwidth]{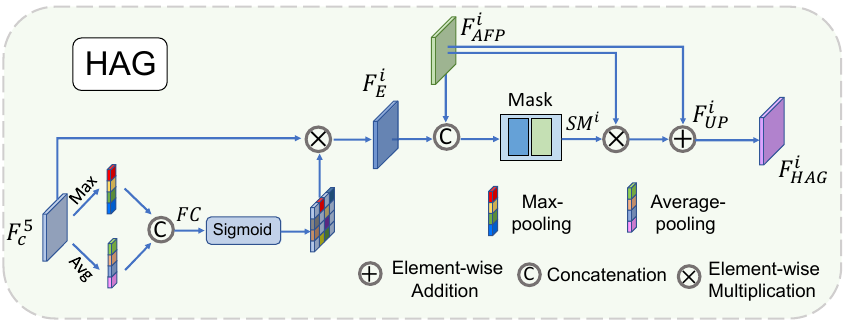}
	\caption{The architecture of High-level Abstract Guidance Module.}
	\label{fig5.2}
	\vspace{-0.15cm}
\end{figure}

As shown in Fig. \ref{fig5.2}, we first enhance the top-layer color features $\mathrm{F}_c^5$ at the spatial level \cite{CBAM}, thereby generating the reweighted color features $\mathrm{F}_E^i$ that highlight the important locations:
\begin{equation}
\mathrm{Act}_{SL}^i=Sigmoid(FC(Concat(Max(\mathrm{F}_c^5),Avg(\mathrm{F}_c^5)))),
\end{equation}

\begin{equation}
\mathrm{F}_E^i=\mathrm{F}_c^5\otimes\mathrm{Act}_{SL}^i,
\end{equation}
where $FC$ denotes the fully connected layers, $Max$ and $Avg$ are the max-pooling operation along channel dimension and global average pooling, respectively, and $\mathrm{Act}_{SL}^i$ is the spatial-level attention.

Considering the auxiliary role of semantic features, we still take the depth reconstruction features as the dominant one in the guiding process. Thus, we concatenate the enhanced color features with the original reconstruction features, and then generate a semantic mask:
\begin{equation}
\mathrm{SM}^i = Conv_{1 \times 1}(PReLU(Conv_{3 \times 3}(Concat(\mathrm{F}_{AFP}^{i},\mathrm{F}_E^i)))),
\end{equation}
where $PReLU$ is the parametric rectified linear unit, and $Conv_{n \times n}$ denotes the convolution layer with the kernel size of $n\times n$.

\begin{table*}[!htbp]
	\setlength{\abovecaptionskip}{0.2cm}
	\setlength{\belowcaptionskip}{-0.1cm}
	\vspace{0.3cm}
    \centering
	\caption{Quantitative DSR results (in MAD $\downarrow$) on the Middlebury 2005 dataset. The best performance is marked in bold, and the second-best performance is underlined.}
	\label{tab:tab1}
	\renewcommand{\arraystretch}{1.1}
	\setlength{\tabcolsep}{1.17mm}{
		\begin{tabular}{c|ccc|ccc|ccc|ccc|ccc|ccc||ccc}
			\hline
			\multirow{2}{*}{Methods} & \multicolumn{3}{c|}{Art} & \multicolumn{3}{c|}{Books} & \multicolumn{3}{c|}{Dolls} & \multicolumn{3}{c|}{Laundry} & \multicolumn{3}{c|}{Mobius} & \multicolumn{3}{c||}{Reindeer} &
			\multicolumn{3}{c}{Average}\\
			\cline{2-22}
			&$\times 4$  &$\times 8$  &  $\times 16$& $\times 4$ & $\times 8$ & $\times 16$ & $\times 4$ & $\times 8$ & $\times 16$ & $\times 4$ & $\times 8$ & $\times 16$ & $\times 4$ & $\times 8$ & $\times 16$ & $\times 4$ & $\times 8$ & $\times 16$ & $\times 4$ & $\times 8$ & $\times 16$ \\
			\hline
			TGV \cite{TGV}& 0.65 & 1.17 & 2.30 & 0.27 & 0.42 & 0.82 & 0.33 & 0.70 & 2.20 & 0.55 & 1.22 & 3.37 & 0.29 & 0.49 & 0.90 & 0.49 & 1.03 & 3.05 & 0.43 & 0.84  & 2.11\\M
			EG \cite{EG}& 0.48 & 0.71 & 1.35 & 0.15 & 0.36 & 0.70 & 0.27 & 0.49 & 0.74 & 0.28 & 0.45 & 0.92 & 0.23 & 0.42 & 0.75 & 0.36 & 0.51 & 0.95 & 0.30 & 0.49 & 0.90\\
			JGF \cite{JGF}& 0.47 & 0.78 & 1.54 & 0.24 & 0.43 & 0.81 & 0.33 & 0.59 & 1.06 & 0.36 & 0.64 & 1.20 & 0.25 & 0.46 & 0.80 & 0.38 & 0.64 & 1.09 & 0.34 & 0.59 & 1.08\\
			CDLLC \cite{CDLLC}& 0.53 & 0.76 & 1.41 & 0.19 & 0.46 & 0.75 & 0.31 & 0.53 & 0.79 & 0.30 & 0.48 & 0.96 & 0.27 & 0.46 & 0.79 & 0.43 & 0.55 & 0.98 & 0.34 & 0.54 & 0.95\\ \hline
			GSRPT \cite{GSRPT}& 0.48 & 0.74 & 1.48 & 0.21 & 0.38 & 0.76 & 0.28 & 0.48 & 0.79 & 0.33 & 0.56 & 1.24 & 0.24 & 0.49 & 0.80 & 0.31 & 0.67 & 1.07 & 0.31 & 0.55 & 1.02\\
			MDDL \cite{MDDL}& 0.46 & 0.62 & 1.87 & 0.24 & 0.37 & 0.73 & 0.29 & 0.51 & 0.79 & 0.32 & 0.53 & 1.11 & 0.19 & 0.37 & 0.74 & 0.41 & 0.51 & 0.95 & 0.32 & 0.49 & 1.03\\
			DEIN \cite{DEIN}& 0.40 & 0.64 & 1.34 & 0.22 & 0.37 & 0.78 & 0.22 & 0.38 & 0.73 & 0.23 &0.36 & 0.81 & 0.20 & 0.35 & 0.73 & 0.26 & 0.40 & 0.80 & 0.26 & 0.42 & 0.87\\
			DJF \cite{DJF}& 0.40 & 1.07 & 2.05 &0.16& 0.45 & 1.00 & 0.20& 0.49 & 0.99 & 0.28 & 0.71 & 1.67 & 0.18 & 0.46& 1.02 & 0.23 & 0.60 & 1.32 & 0.24 & 0.63 & 1.46\\
			CCFN \cite{wen2019}& 0.43 & 0.72 & 1.50 & 0.17 & 0.36 & 0.69 & 0.25 & 0.46 & 0.75 & 0.24 & 0.41 & \textbf{0.71} & 0.23 & 0.39 & 0.73 & 0.29 & 0.46 & 0.95 & 0.27 & 0.47 & 0.89\\
			CTKT \cite{CTKT} & \underline{0.25} & 0.56 & 1.44 & \textbf{0.12} & 0.28 & 0.67 & \underline{0.18} & 0.39 & 0.65 & \underline{0.16} & 0.40 &0.76 & \textbf{0.14} & 0.29 & 0.69 & 0.18 & 0.41 & 0.77 & \underline{0.17} & 0.39 & 0.83\\
			BridgeNet \cite{BridgeNet} & 0.30 & 0.58 & 1.49 & 0.14 & \underline{0.24} & \underline{0.51} & 0.19 & 0.34 & 0.64 & 0.17 & 0.34 &\textbf{0.71} & 0.15 & \underline{0.26} & \underline{0.54} & 0.19 & \underline{0.31} & \underline{0.70} & 0.19 & 0.35 & 0.77\\
			PMBANet \cite{PMBANet}& 0.26 & 0.51 & \underline{1.22} & 0.15 & 0.26 & 0.59 & 0.19 & \underline{0.32} & \underline{0.59} & 0.17 & \underline{0.31} &\textbf{0.71} & 0.16 & \underline{0.26} & 0.67 & \underline{0.17} & 0.33 & 0.74 & 0.18 & \underline{0.33} & \underline{0.75}\\
			MIGNet \cite{MIG} & \textbf{0.21} & \underline{0.47} & \textbf{1.08} & 0.15 & 0.25 & 0.53 & 0.21 & 0.35 & 0.71 & 0.19 & 0.35 &0.81 & 0.15 & \underline{0.26} & 0.55 & 0.22 & 0.36 & 0.82 & 0.19 & 0.34 & \underline{0.75}\\
			HCGNet (Ours)& \textbf{0.21} & \textbf{0.41} & 1.38 & \underline{0.13} & \textbf{0.23} & \textbf{0.44} & \textbf{0.17}  & \textbf{0.31} & \textbf{0.57} & \textbf{0.14} & \textbf{0.28} & 0.78 & \textbf{0.14} & \textbf{0.24} & \textbf{0.45} & \textbf{0.15} & \textbf{0.26} & \textbf{0.64} & \textbf{0.16} & \textbf{0.29} & \textbf{0.71}
			\\
			\hline		
	\end{tabular}}
\end{table*}

\begin{figure*}[]
    \setlength{\abovecaptionskip}{10pt}
	\centering
	\includegraphics[width=0.98\textwidth,
 height=0.45\textwidth]{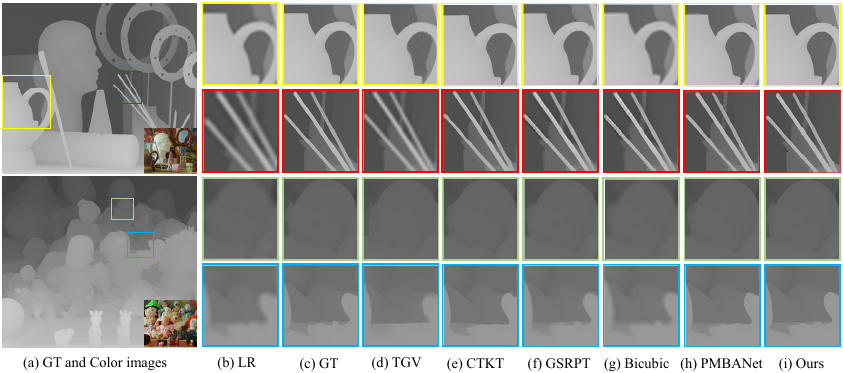}
	\caption{Visual comparisons of $\times$8 upsampling results on the Middlebury 2005 dataset. (a) HR depth maps and color images. (b)-(c) LR depth maps and the HR GT. (d)-(i) The reconstructed HR depth maps by the TGV, CTKT, GSRPT, Bicubic, PMBANet, and Our method, respectively. }
	\label{fig: vis}
\end{figure*}
With the semantic mask, the original depth reconstruction features can be refined by:
\begin{equation}
\mathrm{F}_{HAG}^i = \mathrm{SM}^i \otimes \mathrm{F}_{AFP}^{i} + \mathrm{F}_{AFP}^{i},
\end{equation}
where $\mathrm{F}_{HAG}^i$ are the output of the corresponding HAG module. 
It should be noted that $\mathrm{F}_{HAG}^i$ are the final reconstruction feature at each stage, where the features of the last layer $\mathrm{F}_{HAG}^1$ will be directly used to generate the upsampled depth map, and the reconstruction features of other layers will be sent to the AFP module through the dense transmission to realize the progressive learning of the entire network.


\section{Experiments}
\subsection{Datasets and Implementation Details}
To demonstrate the effectiveness of the proposed method, we conduct comprehensive experiments on the Middlebury dataset, NYU v2 dataset \cite{NYU}, real-world RGB-D-D dataset \cite{FDSR}, and Lu dataset \cite{Lu}. All these datasets are produced with the alignment of color and depth images, and all the ground-truth values of the upsampled depth map are owned by the dataset. From the perspective of dataset construction, for depth cameras such as Kinect, most of the scenes are RGB-D aligned with default parameters. With tools such as Matlab, the accuracy can be further improved by co-calibration. In addition, the device usually includes an SDK that enables users to interpolate the depth output to match the resolution of RGB images. However, the quality of the interpolated depth map can not be guaranteed. Thus, the NYU v2 dataset undergoes additional processing such as depth completion and calibration for improved quality and accuracy.
\begin{itemize}[noitemsep, topsep=0pt]
	\item We collect 36 RGB-D images from Middlebury dataset (6, 21, 9 images from 2001 \cite{middle2001}, 2006 \cite{middle2006}, and 2014 \cite{middle2014} datasets, respectively) for training, and 6 standard depth maps from Middlebury 2005 dataset \cite{middle2005} for testing. 
    The resolution of the image is mostly around $1000\times 1000$.
    \item As for the NYU v2 dataset, it is a real-scene dataset captured by the Kinect camera. We use the first 1000 pairs with the resolution of $640 \times 480$ for training and the remaining 449 pairs for testing. The model trained on the NYU v2 dataset is also directly used to test on the Lu dataset and RGB-D-D dataset for generalization evaluation.
	\item The RGB-D-D dataset is a real-world indoor dataset captured by a Huawei P30 pro cellphone equipped with the color camera and time of flight (TOF) camera, also with a resolution of $640 \times 480$. Following the setting in \cite{FDSR}, 2215 pairs are used for training and 405 pairs for evaluation.
	\item The Lu dataset only consists of 6 RGB-D pairs acquired by the ASUS Xtion Pro camera, all of which are used for testing.
\end{itemize}

\begin{table*}[!t]
 	\renewcommand\arraystretch{1.1}
	\setlength{\abovecaptionskip}{0.2cm}
	\setlength{\belowcaptionskip}{-0.1cm}
	\caption{Quantitative comparisons with state-of-art methods (in RMSE $\downarrow$) on the NYU v2 dataset. The best performance is marked in bold, and the second-best performance is underlined. Note that, the depth values are measured in centimeters}.
	\centering
	\label{tab:tab2}
	\setlength{\tabcolsep}{0.5mm}{
		\begin{tabular}{c|cccccccccccc}
			\hline
			& Bicubic & TGV \cite{TGV} & DJFR \cite{DJFR} &SDF \cite{SDF}  & SVLRM \cite{SVLMR} & DKN \cite{DKN}& FDKN \cite{DKN}
			& FDSR \cite{FDSR}
			& PMBANet \cite{PMBANet}
            & JIIF \cite{acm2021}
            & DCT \cite{22cvpr}
			& HCGNet (Ours)\\
			\hline
			$\times 4$ & 8.16 & 6.98  & 3.38 & 3.04&1.74&1.62&1.86&1.61&\underline{1.30} &1.37 &1.59&\textbf{1.22}\\
			
			$\times 8$ & 14.22 & 11.23  & 5.86 &5.67 &5.59&3.26&3.58&3.18&\underline{2.75} &2.76 &3.16 &\textbf{2.53} \\
			
			$\times 16$ & 22.32 & 28.13  & 10.11 &9.97 &7.23&6.51&6.96&5.86&5.48 &\underline{5.27} &5.84 &\textbf{4.85}\\
			\hline
	\end{tabular}}
\end{table*}

\begin{figure*}[!t]
    \setlength{\abovecaptionskip}{10pt}
	\centering
	\includegraphics[width=1\textwidth]{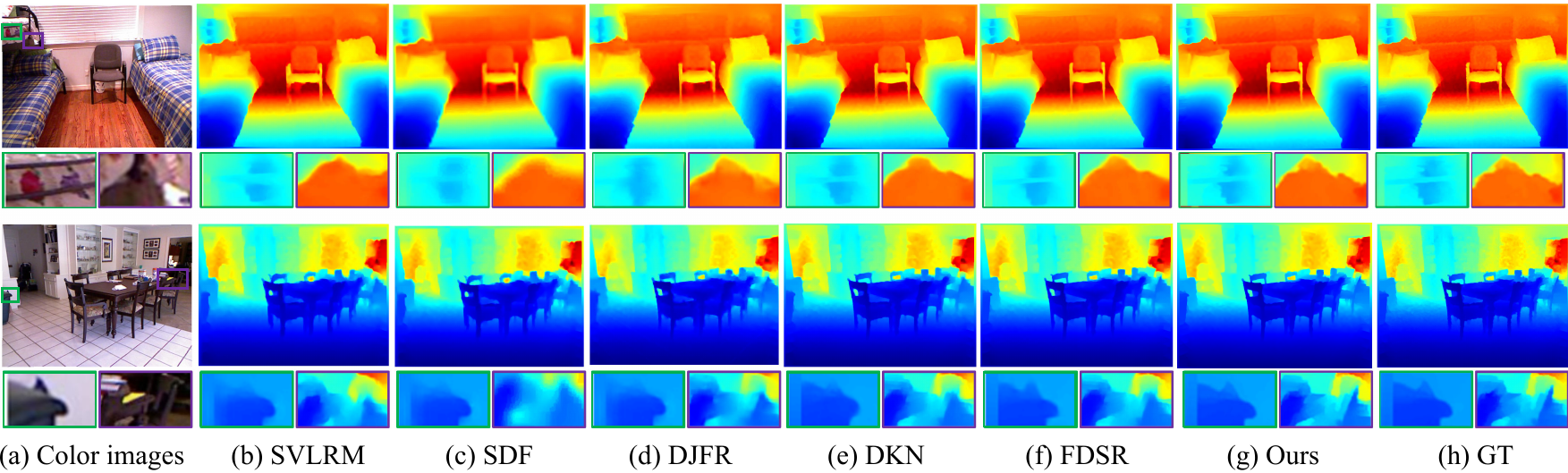}
	\caption{Visual comparisons of different methods under ×8 upsampling on the NYU v2 dataset. (a) Color images. (b) SVLRM. (c) SDF. (d) DJFR. (e) DKN. (f) FDSR. (g) Ours. (h) GT.}
	\label{fig:vis2}
\end{figure*}

For quantitative evaluation, the metrics of Mean Absolute Difference (MAD) and Root Mean Square Error (RMSE) are introduced \cite{BridgeNet,crm/JEI16/underwater,crm/spl21/underwater}. Following PMBANet \cite{PMBANet}, we augment our training samples by cropping the HR pairs into around 15000 HR patches with the squared size of 64, 128, and 256 according to the upsampling factors of $\times4$, $\times8$, and $\times16$. Meanwhile, we perform horizontal and vertical flipping of the training samples with a probability of 0.5. 
The LR depth patches are downsampled into a fixed size of $16\times16$ from HR depth patches via Bicubic interpolation. With respect to the structural details of our HCPNet, in the feature extraction phase, the network maps the RGB-D features to each of the 64-channel dimensions and maintains the number of channels constant throughout the feature extraction phase. For the LDE, HAG, and MCE modules, the input channel dimension changes depending on the number of input features, and the output is uniformly 64 channels. Our HCGNet is implemented by PyTorch with an NVIDIA 3090 GPU. 
As a hyperparameter, the batch size is set to 8 at $\times8$ and $\times4$ factor cases and set to 4 for the $\times16$ factor case.
During training, we use Adam optimizer with momentum of $0.9$, and the network optimization parameters of $\beta_1$, $\beta_2$, and $\epsilon$ are set to $0.9$, $0.99$, and $1e^{-8}$, respectively. The initial learning rate is set to $1e^{-4}$, and it is decreased by multiplying by 0.1 for the first 100 epochs and the next 50 epochs.


\subsection{Performance Comparison}
\textbf{1) Middlebury Dataset:} We compare our method with some state-of-the-art DSR methods under different upsampling factors ($\times 4$, $\times 8$, and $\times 16$), including four traditional depth SR methods (\ie, TGV \cite{TGV}, EG \cite{EG}, JGF \cite{JGF}, and CDLLC \cite{CDLLC}) and nine deep-learning based methods (\ie, GSRPT \cite{GSRPT}, MDDL \cite{MDDL}, DEIN \cite{DEIN}, DJF \cite{DJF}, CCFN \cite{wen2019}, CTKT \cite{CTKT}, BridgeNet\cite{BridgeNet}, PMBANet \cite{PMBANet} and MIGNet \cite{MIG}). 

The quantitative comparisons of the MAD score are reported in Table \ref{tab:tab1}. We can observe that the traditional DSR models often fail to achieve satisfactory performance, especially when dealing with more complex scenes (\eg, Art) or larger upsampling factors (\eg, $\times 16$). 
In contrast, the deep-learning based DSR models show more competitive performance, in which the proposed  HCGNet outperforms other SOTA methods among different scenarios and achieves the best overall average MAD performance under different upsampling factors.
Moreover, our method achieves obvious performance improvement under a challenging large sampling factor (such as $\times 16$). Compared with the \textbf{second best} method, the MAD value of the Mobius scene is decreased from 0.54 to 0.45, with a percentage gain of 16.7\%, and the MAD percentage gain of the Books scene reaches 13.7\%.
Fig. \ref{fig: vis} demonstrates the visual comparisons of different methods under the factor of $\times 8$. As visible, our method can recover more complete and accurate depth details. In the first image, compared with the results of other methods, the boundaries around sticks are sharper and smoother with less jagged noise, and the structure of the distant object is more complete and continuous. 
For the Dolls image, our method can restore the clear and sharp boundaries of the reconstructed objects. For example, the left ear of the doll in the third row is restored with complete contour and the surrounding values are artifact-free, and the sleeves of the doll (\eg, the raised part in the middle) shown in the last row are also recovered more satisfactorily than other methods.
In general, the comparison methods can more or less introduce additional shape distortion and noise, resulting in blurring, discontinuities, and changed depth values in the reconstructed images.
In contrast, our method is able to recover objects more completely while taking into account the restoration of depth details without introducing destructive contamination. 

\begin{table*}[!t]
 	\renewcommand\arraystretch{1.1}
	\setlength{\abovecaptionskip}{0.2cm}
	\setlength{\belowcaptionskip}{-0.1cm}
	\caption{Quantitative comparisons with state-of-art methods (in RMSE $\downarrow$) on the RGB-D-D dataset. The best performance is marked in bold, and the second-best performance is underlined. The depth values are measured in {centimeters}. FDSR$^+$ and HCGNet$^+$ are the models retrained on the RGB-D-D dataset.}
	\centering
	\label{tab:tab3}
	\setlength{\tabcolsep}{2mm}{
		\begin{tabular}{c|cccccccccc}
			\hline
			&SDF \cite{SDF}  & DJFR \cite{DJFR}    & FDKN \cite{DKN} & DKN \cite{DKN}   & FDSR \cite{FDSR} & FDSR$^+$ & DCT \cite{22cvpr} &JIIF \cite{acm2021}
			& HCGNet (Ours) & HCGNet$^+$ (Ours)\\
			\hline
			$\times 4$ &2.00  & 3.35 & 1.18 & 1.30  &1.16&1.11& \underline{1.08} &1.17&1.13&\textbf{0.99}\\
			
			$\times 8$ &3.23 & 5.57 & 1.91 & 1.96 &1.82  &\underline{1.71} &1.74 &1.80 &1.77&\textbf{1.49} \\
			
			$\times 16$ &5.16 & 8.15  & 3.41 & 3.42 &3.06 &3.01 &3.05 &2.84 &\underline{2.70}&\textbf{2.00}\\
			\hline
	\end{tabular}}
\end{table*}

\begin{figure*}[!t]
    \setlength{\abovecaptionskip}{10pt}
	\centering
	\includegraphics[width=1\textwidth]{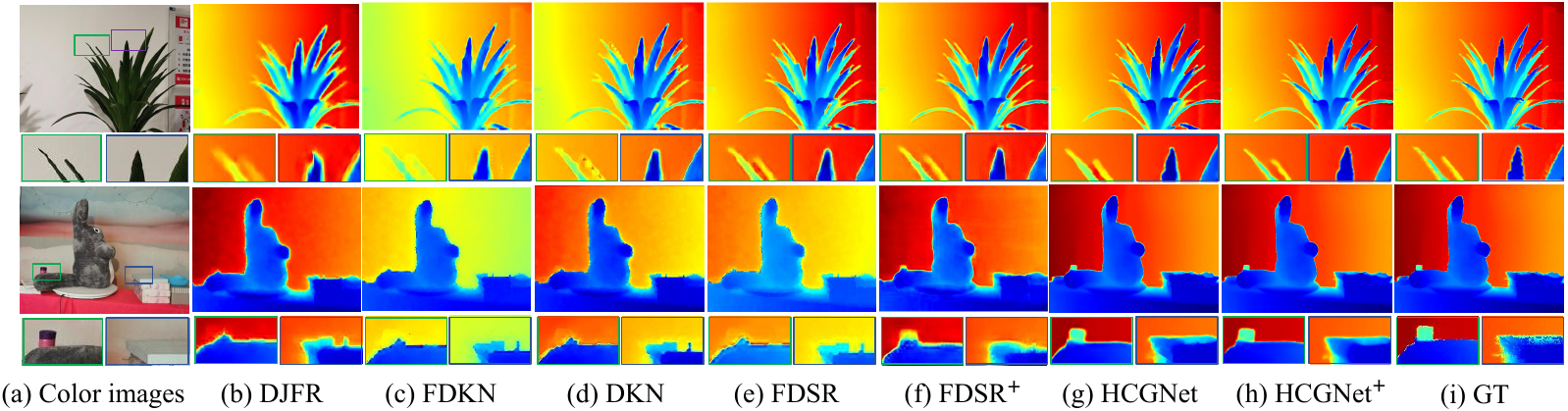}
	\caption{Visual comparisons of different methods under ×8 upsampling on the RGB-D-D dataset. (a) Color images. (b) DJFR. (c) FDKN. (d) DKN. (e) FDSR. (f) FDSR$^+$. (g) HCGNet (Ours). (h) HCGNet$^+$ (Ours). (i) GT.}
	\label{fig:vis4}
\end{figure*}

\textbf{2) NYU v2 Dataset}: We evaluate our method on the NYU v2 dataset and compare it with other SOTA methods, including Bicubic, TGV \cite{TGV}, DJFR \cite{DJFR}, SDF \cite{SDF}, SVLRM \cite{SVLMR}, DKN \cite{DKN}, FDKN \cite{DKN}, FDSR \cite{FDSR}, PMBANet \cite{PMBANet}, JIIF \cite{acm2021} and DCT \cite{22cvpr}.

The quantitative results are provided in Table \ref{tab:tab2}. We can see that our method outperforms all the SOTA methods under different upsampling factors. In the most challenging $\times 16$ upsampling situation, compared with the \textbf{second best} method, the RMSE of our method reaches $4.85$, with a percentage gain of $7.4\%$.
Fig. \ref{fig:vis2} shows some visual comparisons on the NYU v2 dataset under the $\times 8$ factor case. It can be seen that our method has obvious advantages in boundary reconstruction and depth preservation. 
For example, in the overlapping regions of the bed sheet and bed frame marked in purple in the first image, the depth map reconstructed by our method has sharper boundaries and more accurate depth values.
In the second image, our method shows a stronger ability to portray details, such as the object above the trash can in the left image and the oblique border area of the chair in the right image.


\textbf{3) RGB-D-D Dataset}: On this dataset, we evaluate our method (\ie, HCGNet and HCGNet$^+$) and other SOTA methods, including SDF \cite{SDF}, DJFR \cite{DJFR}, DKN \cite{DKN}, FDKN \cite{DKN}, FDSR \cite{FDSR}, FDSR$^+$, DCT \cite{22cvpr}, and  JIIF \cite{acm2021}. The superscript `$+$' indicates the corresponding model was retrained on the RGB-D-D dataset, and no superscript mark indicates the model was only trained on the NYU v2 dataset \cite{NYU} without retraining or fine-tuning.


The quantitative results are reported in Table \ref{tab:tab3}. 
Compared to other models without retraining, our method (\ie, HCGNet) achieves the best performance under large factor cases and even outperforms the retrained FDSR$^+$ method under the most challenging case of $\times16$, which also demonstrates the generalizability of our model.
Concretely, compared with the \textbf{second best} method without retraining, the average RMSE value drops from 3.05 to 2.70 under $\times16$ case with a percentage gain of 11.5\%, and the percentage gain under $\times 8$ factor reaches 5.5\%.
At the same time, we can see a further improvement in the performance of the model after retraining. Our retrained version (\ie, HCGNet$^+$) achieves the best performance under all factor cases and the percentage gain reaches 33.6\% under $\times16$ case compared with the retained FDSR$^+$ method.
For the visual comparison, Fig. \ref{fig:vis4} demonstrates the details on the RGB-D-D dataset under $\times 8$ case. In the first image, our method restores the leaves with clear and sharp boundaries, and even the sawtooth shape of the leaves in the lower right corner is partially restored. In the more challenging second image, our method recovers not only the accurate values both around and inside the cube in the lower left corner but also the complete boundaries of the books.

\textbf{4) Lu Dataset}: We also evaluate our method on the Lu Dataset and compare it with other SOTA methods, including Bicubic, FDKN \cite{DKN}, DKN \cite{DKN}, FDSR \cite{FDSR}, DCT \cite{22cvpr} and JIIF \cite{acm2021}. The Lu dataset was captured in low light and a relatively complex environment, which challenges the generalization ability of the method. The scale of the dataset is relatively small, so all methods are not retrained on this dataset.
As shown in Table \ref{tab:tab4}, our method achieves competitive performance compared with other methods, especially at the large factor cases (\ie, $\times$8, $\times$16). For example, compared with the \textbf{second best} method, the average RMSE value is decreased from 4.16 to 3.75 under $\times 16$ factor case, with a percentage gain of 9.9\%. Fig. \ref{fig:vis3} shows a visual example of different methods on the Lu dataset. As visible, the proposed method punches above its weight in terms of detail reconstruction, such as the sharper borders around the statue and more accurate depth values (\eg, the ears) inside the statue.

\begin{table}[!t]
\setlength{\abovecaptionskip}{0.2cm}
\setlength{\belowcaptionskip}{-0.1cm}
\caption{Quantitative comparisons with state-of-art methods (in RMSE $\downarrow$) on the Lu dataset. The best performance is marked in bold, and the second-best performance is underlined.}
\centering
\label{tab:tab4}
\setlength{\tabcolsep}{0.7mm}{
\begin{tabular}{c|ccccccc}
\hline
 &Bicubic  & FDKN \cite{DKN} & DKN \cite{DKN}  &FDSR \cite{FDSR} &DCT \cite{22cvpr} & JIIF \cite{acm2021} &Ours\\
\hline
$\times 4$ &2.42  &\textbf{0.82} &0.96  &1.29 & 0.88 &\underline{0.85} &1.02\\
$\times 8$ &4.54  &2.10 &2.16   &2.19 & 1.85 &\underline{1.73} &\textbf{1.68}\\
$\times 16$ &7.38  &5.05 &5.11  &5.00 &4.39 &\underline{4.16} &\textbf{3.75}\\
\hline
\end{tabular}}
\end{table}

\begin{figure}[t]
    \setlength{\abovecaptionskip}{8pt}
	\centering
	\includegraphics[width=0.48\textwidth]{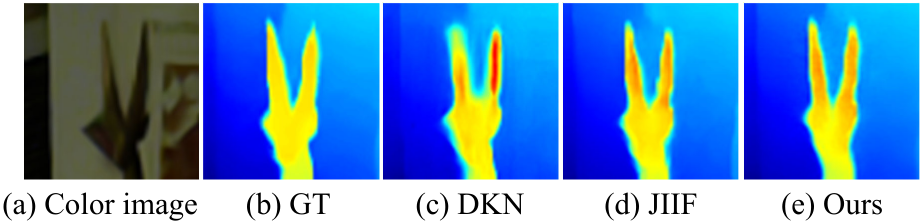}
	\caption{Visual comparisons of different methods under ×8 upsampling on the Lu dataset. (a) Color image. (b) GT. (c) DKN. (d) JIIF. (e) Ours. }
	\label{fig:vis3}
\end{figure}

\subsection{Ablation Study}\label{ab}
\textbf{1) Validation of modules.} Ablation studies are conducted to verify the effectiveness of each key component designed in the proposed HCGNet on both the Middlebury 2005 dataset and the NYU v2 dataset. We remove the color branch and simplify the AFP module as the baseline model. Specifically, we replace the MCE block with a few simple convolution layers and replace the AAP block with the residual block accordingly. Then, we sequentially add the HAG, LDE, and AFP modules to the baseline to verify the effectiveness of each module. For the validation of MCE and AAP blocks in AFP, we replace the MCE with the similar RDB \cite{RDB} block and the AAP module with the similar DBPN \cite{DBPN} block. The quantitative results under the $\times 8$ case are reported in Table \ref{tab:tab5}, and some visual examples are shown in Fig. \ref{fig:ab1}.




\begin{table}[!t]
\setlength{\abovecaptionskip}{0.2cm}
\setlength{\belowcaptionskip}{-0.1cm}
\caption{Ablation studies of our HCGNet in terms of average MAD ($\downarrow$) values on the Middlebury 2005 dataset and RMSE ($\downarrow$) values on the NYU v2 dataset ($\times 8$ case).}
\centering
\label{tab:tab5}
\setlength{\tabcolsep}{1mm}{
\begin{tabular}{c|ccccc|c|c}
\hline
Model &Baseline & HAG & LDE & RDB+DBPN & AFP & MAD & RMSE \\
\hline
1 &\checkmark & & & &  & 0.42 &3.80 \\
\hline
2&\checkmark & \checkmark & & & & 0.36 &2.88 \\
3&\checkmark & \checkmark  &\checkmark & & & 0.34 &2.74 \\
4&\checkmark & \checkmark  &\checkmark &\checkmark &  &0.32 &2.61 \\
\hline
5&\checkmark & \checkmark  &\checkmark &  &\checkmark  &\textbf{0.29} &\textbf{2.53} \\
\hline
\end{tabular}}
\end{table}

\begin{table}[!t]
\setlength{\abovecaptionskip}{0.2cm}
\setlength{\belowcaptionskip}{-0.1cm}
\caption{Quantitative evaluation in RMSE ($\downarrow$) with different number of AFP and AAP modules on the NYU v2 dataset ($\times 8$ case).}
\centering
\label{tab:tab6}
\setlength{\tabcolsep}{3.5mm}{
\begin{tabular}{c|c||c|c}
\hline
Model &RMSE& Model & RMSE  \\
\hline
5AFPs (Ours) &\textbf{2.53} &4AAPs (Ours) &\textbf{2.53} \\
\hline
4AFPs &2.63 & 3AAPs &2.64 \\
3AFPs &2.68 & 5AAPs  &2.61 \\
\hline
\end{tabular}}
\end{table}

First, we verify the role of color guidance in DSR. After introducing the HAG module into the baseline to provide high-level abstract guidance, we can see that the MAD value on the Middlebury dataset is improved from 0.42 to 0.36 with the percentage gain of 14.3\%, and RMSE values on the NYU v2 dataset decreased from 3.80 to 2.88 with the percentage gain of 24.2\%. 
As visible, compared with the second and third columns in the first image of Fig. \ref{fig:ab1}, the indistinguishable hole regions in the baseline model (marked in red box) have been obviously restored. Then, we embed the color detail information into the depth reconstruction branch through the LDE module. As reported in model 3 of Table \ref{tab:tab5}, the performance is further improved, where the percentage gains of the MAD value on the Middlebury 2005 dataset and RMSE value on the NYU v2 dataset reach 5.6\% and 4.9\% compared with model 2, respectively. At the same time, as shown in Fig. \ref{fig:ab1}(d), the holes in the visualization result are also reconstructed more clearly. In summary, all these results demonstrate that our proposed HAG and LDE modules provide effective guidance information for depth map super-resolution reconstruction from two perspectives, and improve the reconstruction accuracy.

\begin{figure}[!t]
    \setlength{\abovecaptionskip}{1pt}
	\centering
	\includegraphics[width=0.48\textwidth]{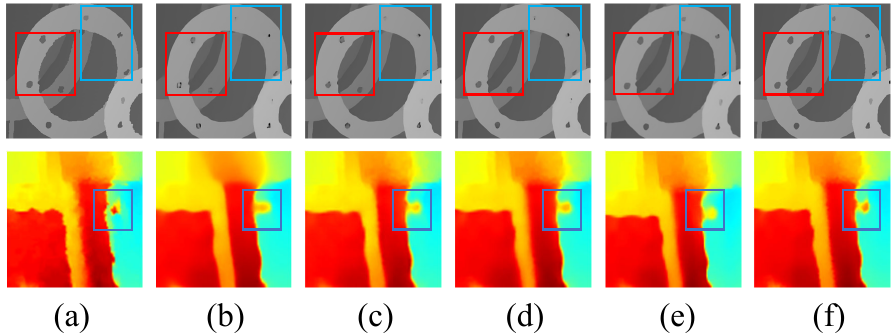}
	\caption{Visual comparisons of different ablation models (×8 case). (a) Ground truth. (b) Baseline (model 1). (c) Baseline + HAG (model 2). (d) Baseline + HAG + LDE (model 3). (e) Baseline + HAG + LDE + RDB + DBPN (model 4). (f) Baseline + HAG + LDE + AFP (model 5). }
	\label{fig:ab1}
\end{figure}

Furthermore, we verify the effectiveness of the proposed AFP module. On the basis of our DSR framework for color guidance, we introduce the AFP module to achieve multi-scale content enhancement and adaptive attention projection. Compared with model 3 in Table \ref{tab:tab5}, the MAD score is further improved from 0.34 to 0.29 with a percentage gain of 14.7\%, and the RMSE score is decreased from 2.74 to 2.53 with a percentage gain of 7.7\%. In terms of the visualization results shown in Fig.  \ref{fig:ab1}(f), the final full model (model 5) can not only maintain the details of the holes in the red-marked area but also effectively update the incorrectly reconstructed holes in the lower right corner of model 3 in the first row. To further illustrate the superiority of the AFP module design over similar methods, we add the ablation experiments of model 4. Comparing model 4 and model 5, we can see that after replacing MCE and AAP with RDB and DBPN respectively, the model performance drops obviously. On the Middlebury 2005 dataset, the MAD score increases from 0.29 to 0.32, with a performance drop of 10.3\%. While on the NYU v2 dataset, the RMSE score increases from 2.53 to 2.61, with a performance drop of 3.2\%. 
The experimental results also prove the effectiveness of our proposed AFP module in-depth reconstruction.

Finally, we conduct an ablation study to compare the performance of different numbers of modules within the AFP and AAP components, the results of which are illustrated in Table \ref{tab:tab6}. Specifically, we conduct experiments on the AFP modules by sequentially removing them from level 5 and level 4 of the network, while still retaining the HAG’s global guidance. The results indicate that the removal of one AFP module (reducing the number from five to four) leads to an increase in RMSE from 2.53 to 2.63, corresponding to a performance decrease of 4\%. Since the proposed HCGNet is a hierarchical progressive reconstruction, the quality of the reconstruction at each layer has an essential impact on subsequent layers. For the ablation of AAP blocks, compared with the four AAP blocks, the RMSE value of three blocks increases from 2.53 to 2.64, with a performance drop of 4.3\%. Meanwhile, observation reveals that too many AAP blocks can also lead to performance degradation. This is because, at the current stage, the quality of the depth features is limited, and the correction capability of the AAP module is similarly constrained. Excessively deep stacking may, in fact, hinder the restoration of the depth features. Based on the above experimental results, we finally set the number of AAP to 4.

\begin{figure}[!t]
    \setlength{\abovecaptionskip}{10pt}
	\centering
	\includegraphics[width=0.48\textwidth]{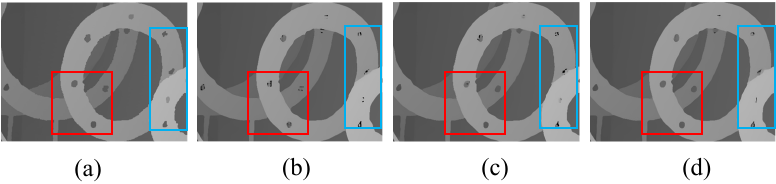}
	\caption{Visual examples of adding our modules to PMBANet. (a) GT. (b) PMBANet$^*$. (c) PMBANet$^*$$+$LDE. (d) PMBANet$^*$$+$LDE$+$HAG. }
	\label{fig: pmba-color}
\end{figure}

\textbf{2) Portability of modules.}
To further validate the effectiveness of the proposed LDE and HAG modules, we conduct validation experiments by adding LDE and HAG modules to the PMBANet \cite{PMBANet}, which is a network considering only low-level color information (Mode (a) in Fig. \ref{fig1}). Specifically, we use the depth features of the corresponding layer, which is modified by the LDE module, as the input of the current MBA block, while the global HAG module is used to refine the depth features after each reconstruction branch. As shown in Table \ref{tab: pmba-nyu}, it can be seen that after adding these two modules to the PMBANet, the performance is improved on both two datasets. For example, on the Middlebury 2005 dataset, with both LDE and HAG modules, compared with the original PMBANet, the average MAD value is improved from 0.33 to 0.30. 
On the NYU v2 dataset, introducing both the LDE and HAG modules reduces the RMSE of PMBANet from 2.75 to 2.58 with a percentage gain of 6.2\%.  
Some visual examples of adding our modules to PMBANet are shown in Fig. \ref{fig: pmba-color}. We can see that, compared with the original result of PMBANet, the PMBANet model with our LDE and HAG modules improves the accuracy and details of the reconstruction, such as the hole regions. 
These experiments all again demonstrate the effectiveness and portability of our proposed hierarchical color guidance mechanism.

\begin{table}[!t]
\centering

\caption{Quantitative evaluation of adding our modules to PMBANet, including the MAD ($\downarrow$) on the Middlebury 2005 dataset and the RMSE ($\downarrow$) on the NYU v2 dataset ($\times8$ case). PMBANet$^*$ denotes the retrained version.}
\label{tab: pmba-nyu}
\setlength{\tabcolsep}{2.7mm}{
\begin{tabular}{c|c|c|c}
			\hline
			\multirow{1}{*}{Methods} & \multicolumn{1}{c|}{MAD} & \multicolumn{1}{c|}{RMSE} & \multicolumn{1}{c}{Running Time}  \\
			\cline{1-3}        
			\hline
			PMBANet$^*$ \cite{PMBANet} &0.33 & 2.75& \textbf{13.6ms} \\
            PMBANet$^*$$+$LDE &0.31 & 2.72& 18.8ms \\
            PMBANet$^*$$+$LDE$+$HAG &\textbf{0.30} & \textbf{2.58}& 21.1ms \\
\hline
\end{tabular}}
\end{table}

\subsection{Discussion}
For the training time, it takes about 24 hours to obtain the final $\times8$ model for 200 epochs with a batch size of 8. And our $\times4$ model only requires about 10 hours for training with a batch size of 8. This suggests that the training time is correlated with both our specific task and batch size settings, thus existing methods focus more on the running time of the test.
The running time comparisons on the NYU v2 dataset ($640\times 480$) are shown in Table \ref{tab:tab8}. It can be seen that our method processes $\times8$ DSR task in less than 20ms per image on the NYU v2 dataset ($640\times 480$) while achieving optimal performance compared to other SOTA methods. In other words, our model achieves a good balance in terms of performance and efficiency.

\begin{figure}[!t]
    \setlength{\abovecaptionskip}{10pt}
	\centering
    \includegraphics[width=0.4\textwidth]
    {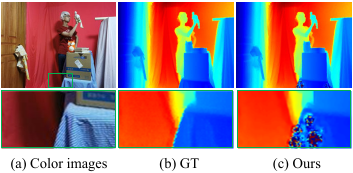}
	\caption{Illustration of the failure cases under the $\times4$ factor on the RGB-D-D dataset.}
	\label{fig: failure cases}
\end{figure}

In terms of performance, the advantages are more pronounced when dealing with more complex scenes or larger upsampling factors. 
But as reported in Table \ref{tab:tab3} and Table \ref{tab:tab4}, our method has no obvious advantages when dealing with low-scale ($\times4$) DSR on the Lu dataset and real-world RGB-D-D dataset. Some failure cases under the $\times4$ factor on the RGB-D-D dataset are also provided in Fig. \ref{fig: failure cases}. It has been observed that our algorithm may suffer from reconstruction errors when confronted with regions of sudden changes in brightness inside objects in the low-depth range of the scene. In these regions, the brightness boundary of the color image varies greatly, but the depth variation is small, so the inconsistency between the color brightness and depth boundaries may cause reconstruction errors in small regions. Therefore, the robustness of the model to such challenging scenarios needs to be further optimized in the future.

\begin{table}[!t]
\centering
\caption{Average running time and RMSE ($\downarrow$) of different methods on the NYU v2 dataset ($\times8$ case).}
\label{tab:tab8}
\setlength{\tabcolsep}{4mm}{
\begin{tabular}{c|c|c}
			\hline
			\multirow{1}{*}{Methods} & \multicolumn{1}{c|}{RMSE} & \multicolumn{1}{c}{Running Time}  \\
			\cline{1-3}
			\hline
			PMBANet$^*$ (TIP'20) \cite{PMBANet}& 2.75& \textbf{13.6ms} \\
            DKN (IJCV'21) \cite{DKN} & 3.26& 81.1ms \\
            JIIF (ACMMM'21) \cite{acm2021} & 2.76& 132.0ms \\
            DCT (CVPR'22) \cite{22cvpr} & 3.21& 46.5ms \\
            Ours & \textbf{2.53}& 19.5ms \\            
\hline
\end{tabular}}
\end{table}

\section{Conclusion}
In this paper, we rethink the utilization of color information in DSR and propose a novel framework HCGNet. On the one hand, to supplement the high-frequency color information for the depth features, we embed the low-level detail features through the LDE module; On the other hand, to maintain semantic consistency in the reconstruction process, we encode the global abstract guidance information in the HAG module. In addition, we design the AFP module to make full use of multi-scale information while projecting effective information for reconstruction in an attention manner. Experiments on four benchmark datasets demonstrate that the proposed network outperforms other state-of-the-art methods both qualitatively and quantitatively.


\par
\ifCLASSOPTIONcaptionsoff
  \newpage
\fi
{
\bibliographystyle{IEEEtran}
\bibliography{ref}

\begin{thebibliography}{10}
\providecommand{\url}[1]{#1}
\csname url@samestyle\endcsname
\providecommand{\newblock}{\relax}
\providecommand{\bibinfo}[2]{#2}
\providecommand{\BIBentrySTDinterwordspacing}{\spaceskip=0pt\relax}
\providecommand{\BIBentryALTinterwordstretchfactor}{4}
\providecommand{\BIBentryALTinterwordspacing}{\spaceskip=\fontdimen2\font plus
\BIBentryALTinterwordstretchfactor\fontdimen3\font minus \fontdimen4\font\relax}
\providecommand{\BIBforeignlanguage}[2]{{%
\expandafter\ifx\csname l@#1\endcsname\relax
\typeout{** WARNING: IEEEtran.bst: No hyphenation pattern has been}%
\typeout{** loaded for the language `#1'. Using the pattern for}%
\typeout{** the default language instead.}%
\else
\language=\csname l@#1\endcsname
\fi
#2}}
\providecommand{\BIBdecl}{\relax}
\BIBdecl

\bibitem{TIM-drive}
H.~Hu, T.~Zhao, Q.~Wang, F.~Gao, L.~He, and Z.~Gao, ``Monocular {3-D} vehicle detection using a cascade network for autonomous driving,'' \emph{{IEEE} Trans. Instrum. Meas.}, vol.~70, pp. 1--13, 2021.

\bibitem{TIM-3D}
Z.~Pan, J.~Hou, and L.~Yu, ``Optimization {RGB-D} {3D} reconstruction algorithm based on dynamic {SLAM},'' \emph{{IEEE} Trans. Instrum. Meas.}, vol.~72, pp. 1--13, 2023.

\bibitem{TIM-3D2}
C.~Gu, Y.~Cong, and G.~Sun, ``Three birds, {O}ne stone: Unified laser-based {3D} reconstruction across different media,'' \emph{{IEEE} Trans. Instrum. Meas.}, vol.~70, pp. 1--12, 2021.

\bibitem{TIM-recognition}
S.~Liu, R.~Wu, J.~Qu, and Y.~Li, ``{HDA-N}et: Hybrid convolutional neural networks for small objects recognization at airports,'' \emph{{IEEE} Trans. Instrum. Meas.}, vol.~71, pp. 1--14, 2022.

\bibitem{TIM-recognition2}
A.~Kosuge, S.~Suehiro, M.~Hamada, and T.~Kuroda, ``mm{W}ave-{YOLO}: {A} mmwave imaging radar-based real-time multiclass object recognition system for {ADAS} applications,'' \emph{{IEEE} Trans. Instrum. Meas.}, vol.~71, pp. 1--10, 2022.

\bibitem{SOD-23-CSVT}
R.~Cong, Q.~Qin, C.~Zhang, Q.~Jiang, S.~Wang, Y.~Zhao, and S.~Kwong, ``A weakly supervised learning framework for salient object detection via hybrid labels,'' \emph{{IEEE} Trans. Circuits Syst. Video Technol.}, vol.~33, no.~2, pp. 534--548, 2023.

\bibitem{crm/tip22/CIRNet}
R.~Cong, Q.~Lin, C.~Zhang, C.~Li, X.~Cao, Q.~Huang, and Y.~Zhao, ``{CIR-Net}: Cross-modality interaction and refinement for {RGB-D} salient object detection,'' \emph{IEEE Trans. Image Process.}, vol.~31, pp. 6800--6815, 2022.

\bibitem{crm/tip21/DynamicRGBDSOD}
H.~Wen, C.~Yan, X.~Zhou, R.~Cong, Y.~Sun, B.~Zheng, J.~Zhang, Y.~Bao, and G.~Ding, ``Dynamic selective network for {RGB-D} salient object detection,'' \emph{{IEEE} Trans. Image Process.}, vol.~30, pp. 9179--9192, 2021.

\bibitem{crm/acmmm21/CDINet}
C.~Zhang, R.~Cong, Q.~Lin, L.~Ma, F.~Li, Y.~Zhao, and S.~Kwong, ``Cross-modality discrepant interaction network for {RGB-D} salient object detection,'' in \emph{Proc. ACM Int. Conf. Multim. (ACM MM)}, 2021, pp. 2094--2102.

\bibitem{crm/ACMMM23/PICRNet}
R.~Cong, H.~Liu, C.~Zhang, W.~Zhang, F.~Zheng, R.~Song, and S.~Kwong, ``Point-aware interaction and {CNN}-induced refinement network for {RGB-D} salient object detection,'' in \emph{Proc. ACM Int. Conf. Multim. (ACM MM)}, 2023, pp. 406--416.

\bibitem{crm/tmm22/3DSaliency}
Y.~Mao, Q.~Jiang, R.~Cong, W.~Gao, F.~Shao, and S.~Kwong, ``Cross-modality fusion and progressive integration network for saliency prediction on stereoscopic {3D} images,'' \emph{{IEEE} Trans. Multimedia}, vol.~24, pp. 2435--2448, 2022.

\bibitem{SOD-22-TGRS}
R.~Cong, Y.~Zhang, L.~Fang, J.~Li, Y.~Zhao, and S.~Kwong, ``{RRN}et: Relational reasoning network with parallel multiscale attention for salient object detection in optical remote sensing images,'' \emph{{IEEE} Trans. Geosci. Remote. Sens.}, vol.~60, pp. 1--11, 2022.

\bibitem{crm/tmm22/TNet}
R.~Cong, K.~Zhang, C.~Zhang, F.~Zheng, Y.~Zhao, Q.~Huang, and S.~Kwong, ``Does {Thermal} really always matter for {RGB-T} salient object detection?'' \emph{IEEE Trans. Multimedia}, early access, doi: 10.1109/TMM.2022.3216476.

\bibitem{crm/tcyb21/ASIFNet}
C.~Li, R.~Cong, S.~Kwong, J.~Hou, H.~Fu, G.~Zhu, D.~Zhang, and Q.~Huang, ``{ASIF-Net}: Attention steered interweave fusion network for {RGB-D} salient object detection,'' \emph{IEEE Trans. Cybern.}, vol.~50, no.~1, pp. 88--100, 2021.

\bibitem{crm/eccv20/RGBDSOD}
C.~Li, R.~Cong, Y.~Piao, Q.~Xu, and C.~C. Loy, ``{RGB-D} salient object detection with cross-modality modulation and selection,'' in \emph{Proc. European Conf. Comput. Vis. (ECCV)}, 2020, pp. 225--241.

\bibitem{cvpr2020}
X.~Song, Y.~Dai, D.~Zhou, L.~Liu, W.~Li, H.~Li, and R.~Yang, ``Channel attention based iterative residual learning for depth map super-resolution,'' in \emph{Proc. IEEE Conf. Comput. Vis. Pattern Recognit. (CVPR)}, 2020, pp. 5630--5639.

\bibitem{WAFP}
X.~Song, D.~Zhou, W.~Li, Y.~Dai, L.~Liu, H.~Li, R.~Yang, and L.~Zhang, ``{WAFP-N}et: Weighted attention fusion based progressive residual learning for depth map super-resolution,'' \emph{{IEEE} Trans. Multim.}, vol.~24, pp. 4113--4127, 2022.

\bibitem{crm/access17/dsr}
M.~Ni, J.~Lei, R.~Cong, K.~Zheng, B.~Peng, and X.~Fan, ``Color-guided depth map super resolution using convolutional neural network,'' \emph{{IEEE} Access}, vol.~5, pp. 26\,666--26\,672, 2017.

\bibitem{DKN}
B.~Kim, J.~Ponce, and B.~Ham, ``Deformable kernel networks for joint image filtering,'' \emph{Int. J. Comput. Vis.}, vol. 129, no.~2, pp. 579--600, 2021.

\bibitem{PDR}
P.~Liu, Z.~Zhang, Z.~Meng, N.~Gao, and C.~Wang, ``{PDR-N}et: Progressive depth reconstruction network for color guided depth map super-resolution,'' \emph{Neurocomputing}, vol. 479, pp. 75--88, 2022.

\bibitem{zuo2020}
Y.~Zuo, Q.~Wu, Y.~Fang, P.~An, L.~Huang, and Z.~Chen, ``Multi-scale frequency reconstruction for guided depth map super-resolution via deep residual network,'' \emph{{IEEE} Trans. Circuits Syst. Video Technol.}, vol.~30, no.~2, pp. 297--306, 2020.

\bibitem{BridgeNet}
Q.~Tang, R.~Cong, R.~Sheng, L.~He, D.~Zhang, Y.~Zhao, and S.~Kwong, ``{B}ridge{N}et: {A} joint learning network of depth map super-resolution and monocular depth estimation,'' in \emph{Proc. ACM Int. Conf. Multim. (ACM MM)}, 2021, pp. 2148--2157.

\bibitem{Simultaneous-RGBD-SR}
L.~Zhao, H.~Bai, J.~Liang, B.~Zeng, A.~Wang, and Y.~Zhao, ``Simultaneous color-depth super-resolution with conditional generative adversarial networks,'' \emph{Pattern Recognit.}, vol.~88, pp. 356--369, 2019.

\bibitem{PMBANet}
X.~Ye, B.~Sun, Z.~Wang, J.~Yang, R.~Xu, H.~Li, and B.~Li, ``{PMBANet}: Progressive multi-branch aggregation network for scene depth super-resolution,'' \emph{{IEEE} Trans. Image Process.}, vol.~29, pp. 7427--7442, 2020.

\bibitem{guo2019}
C.~Guo, C.~Li, J.~Guo, R.~Cong, H.~Fu, and P.~Han, ``Hierarchical features driven residual learning for depth map super-resolution,'' \emph{{IEEE} Trans. Image Process.}, vol.~28, no.~5, pp. 2545--2557, 2019.

\bibitem{DAEA}
X.~Cao, Y.~Luo, X.~Zhu, L.~Zhang, Y.~Xu, H.~Shen, T.~Wang, and Q.~Feng, ``{DAEAN}et: Dual auto-encoder attention network for depth map super-resolution,'' \emph{Neurocomputing}, vol. 454, pp. 350--360, 2021.

\bibitem{2020SOD}
D.~Fan, Y.~Zhai, A.~Borji, J.~Yang, and L.~Shao, ``{BBS}-{N}et: {RGB-D} salient object detection with a bifurcated backbone strategy network,'' in \emph{Proc. Eur. Conf. Comput. Vis. (ECCV)}, 2020, pp. 275--292.

\bibitem{wen2019}
Y.~Wen, B.~Sheng, P.~Li, W.~Lin, and D.~D. Feng, ``Deep color guided coarse-to-fine convolutional network cascade for depth image super-resolution,'' \emph{{IEEE} Trans. Image Process.}, vol.~28, no.~2, pp. 994--1006, 2019.

\bibitem{Yang2007}
Q.~Yang, R.~Yang, J.~Davis, and D.~Nist{\'{e}}r, ``Spatial-depth super resolution for range images,'' in \emph{Proc. IEEE Conf. Comput. Vis. Pattern Recognit. (CVPR)}, 2007, pp. 1--8.

\bibitem{DBPN}
M.~Haris, G.~Shakhnarovich, and N.~Ukita, ``Deep back-projection networks for super-resolution,'' in \emph{Proc. IEEE Conf. Comput. Vis. Pattern Recognit. (CVPR)}, 2018, pp. 1664--1673.

\bibitem{TIM-DEWRN}
W.~Hsu and P.~Jian, ``Detail-enhanced wavelet residual network for single image super-resolution,'' \emph{{IEEE} Trans. Instrum. Meas.}, vol.~71, pp. 1--13, 2022.

\bibitem{TIM-SRN}
S.~Ahmadi, L.~K{\"{a}}stner, J.~C. Hauffen, P.~Jung, and M.~Ziegler, ``Photothermal-{SR-N}et: {A} customized deep unfolding neural network for photothermal super resolution imaging,'' \emph{{IEEE} Trans. Instrum. Meas.}, vol.~71, pp. 1--9, 2022.

\bibitem{zhang2023controlvideo}
Y.~Zhang, Y.~Wei, D.~Jiang, X.~Zhang, W.~Zuo, and Q.~Tian, ``Controlvideo: Training-free controllable text-to-video generation,'' \emph{arXiv preprint arXiv:2305.13077}, 2023.

\bibitem{TIM-SADN}
W.~Shi, F.~Tao, and Y.~Wen, ``Structure-aware deep networks and pixel-level generative adversarial training for single image super-resolution,'' \emph{{IEEE} Trans. Instrum. Meas.}, vol.~72, pp. 1--14, 2023.

\bibitem{TIM-DTSR}
Y.~Zhu, S.~Wang, Y.~Zhang, Z.~He, and Q.~Wang, ``A dual transformer super-resolution network for improving the definition of vibration image,'' \emph{{IEEE} Trans. Instrum. Meas.}, vol.~72, pp. 1--12, 2023.

\bibitem{mvp1}
J.~Wu, R.~Cong, L.~Fang, C.~Guo, B.~Zhang, and P.~Ghamisi, ``Unpaired remote sensing image super-resolution with content-preserving weak supervision neural network,'' \emph{Sci. China Inf. Sci.}, vol.~66, no.~1, 2023.

\bibitem{mvp2}
F.~Li, Y.~Wu, H.~Bai, W.~Lin, R.~Cong, and Y.~Zhao, ``Learning detail-structure alternative optimization for blind super-resolution,'' \emph{{IEEE} Trans. Multim.}, vol.~25, pp. 2825--2838, 2023.

\bibitem{ATGVNet}
G.~Riegler, M.~R{\"{u}}ther, and H.~Bischof, ``{ATGV}-{Net}: Accurate depth super-resolution,'' in \emph{Proc. Eur. Conf. Comput. Vis. (ECCV)}, 2016, pp. 268--284.

\bibitem{song2019}
X.~Song, Y.~Dai, and X.~Qin, ``Deeply supervised depth map super-resolution as novel view synthesis,'' \emph{{IEEE} Trans. Circuits Syst. Video Technol.}, vol.~29, no.~8, pp. 2323--2336, 2019.

\bibitem{sun2020acm}
X.~Ye, B.~Sun, Z.~Wang, J.~Yang, R.~Xu, H.~Li, and B.~Li, ``Depth super-resolution via deep controllable slicing network,'' in \emph{Proc. ACM Int. Conf. Multim. (ACM MM)}, 2020, pp. 1809--1818.

\bibitem{kopf}
J.~Kopf, M.~F. Cohen, D.~Lischinski, and M.~Uyttendaele, ``Joint bilateral upsampling,'' \emph{{ACM} Trans. Graph.}, vol.~26, no.~3, p.~96, 2007.

\bibitem{He2013}
K.~He, J.~Sun, and X.~Tang, ``Guided image filtering,'' \emph{{IEEE} Trans. Pattern Anal. Mach. Intell.}, vol.~35, no.~6, pp. 1397--1409, 2013.

\bibitem{DNR}
J.~Wang, L.~Sun, R.~Xiong, Y.~Shi, Q.~Zhu, and B.~Yin, ``Depth map super-resolution based on dual normal-depth regularization and graph laplacian prior,'' \emph{{IEEE} Trans. Circuits Syst. Video Technol.}, vol.~32, no.~6, pp. 3304--3318, 2022.

\bibitem{huang2019}
L.~Huang, J.~Zhang, Y.~Zuo, and Q.~Wu, ``Pyramid-structured depth {MAP} super-resolution based on deep dense-residual network,'' \emph{{IEEE} Signal Process. Lett.}, vol.~26, no.~12, pp. 1723--1727, 2019.

\bibitem{dilated}
F.~Yu, V.~Koltun, and T.~A. Funkhouser, ``Dilated residual networks,'' in \emph{Proc. IEEE Conf. Comput. Vis. Pattern Recognit. (CVPR)}, 2017, pp. 636--644.

\bibitem{crm/tip20/MCMT-GAN}
Y.~Huang, F.~Zheng, R.~Cong, W.~Huang, M.~R. Scott, and L.~Shao, ``{MCMT-GAN:} multi-task coherent modality transferable {GAN} for {3D} brain image synthesis,'' \emph{{IEEE} Trans. Image Process.}, vol.~29, pp. 8187--8198, 2020.

\bibitem{CBAM}
S.~Woo, J.~Park, J.~Lee, and I.~S. Kweon, ``{CBAM}: Convolutional block attention module,'' in \emph{Proc. Eur. Conf. Comput. Vis. (ECCV)}, 2018, pp. 3--19.

\bibitem{TGV}
D.~Ferstl, C.~Reinbacher, R.~Ranftl, M.~R{\"{u}}ther, and H.~Bischof, ``Image guided depth upsampling using anisotropic total generalized variation,'' in \emph{Proc. IEEE Int. Conf. Comput. Vis. (ICCV)}, 2013, pp. 993--1000.

\bibitem{EG}
J.~Xie, R.~S. Feris, and M.~Sun, ``Edge-guided single depth image super resolution,'' \emph{{IEEE} Trans. Image Process.}, vol.~25, no.~1, pp. 428--438, 2016.

\bibitem{JGF}
M.~Liu, O.~Tuzel, and Y.~Taguchi, ``Joint geodesic upsampling of depth images,'' in \emph{Proc. IEEE Conf. Comput. Vis. Pattern Recognit. (CVPR)}, 2013, pp. 169--176.

\bibitem{CDLLC}
J.~Xie, C.~Chou, R.~S. Feris, and M.~Sun, ``Single depth image super resolution and denoising via coupled dictionary learning with local constraints and shock filtering,'' in \emph{Proc. IEEE Int. Conf. Multimedia Expo (ICME)}, 2014, pp. 1--6.

\bibitem{GSRPT}
R.~de~Lutio, S.~D'Aronco, J.~D. Wegner, and K.~Schindler, ``Guided super-resolution as pixel-to-pixel transformation,'' in \emph{Proc. IEEE Int. Conf. Comput. Vis. (ICCV)}, 2019, pp. 8828--8836.

\bibitem{MDDL}
J.~Wang, W.~Xu, J.~Cai, Q.~Zhu, Y.~Shi, and B.~Yin, ``Multi-direction dictionary learning based depth map super-resolution with autoregressive modeling,'' \emph{{IEEE} Trans. Multim.}, vol.~22, no.~6, pp. 1470--1484, 2020.

\bibitem{DEIN}
X.~Ye, X.~Duan, and H.~Li, ``Depth super-resolution with deep edge-inference network and edge-guided depth filling,'' in \emph{Proc. IEEE Int. Conf. Acoust., Speech Signal Process. (ICASSP)}, 2018, pp. 1398--1402.

\bibitem{DJF}
Y.~Li, J.~Huang, N.~Ahuja, and M.~Yang, ``Deep joint image filtering,'' in \emph{Proc. Eur. Conf. Comput. Vis. (ECCV)}, 2016, pp. 154--169.

\bibitem{CTKT}
B.~Sun, X.~Ye, B.~Li, H.~Li, Z.~Wang, and R.~Xu, ``Learning scene structure guidance via cross-task knowledge transfer for single depth super-resolution,'' in \emph{Proc. IEEE Conf. Comput. Vis. Pattern Recognit. (CVPR)}, 2021, pp. 7792--7801.

\bibitem{MIG}
Y.~Zuo, H.~Wang, Y.~Fang, X.~Huang, X.~Shang, and Q.~Wu, ``{MIG-N}et: Multi-scale network alternatively guided by intensity and gradient features for depth map super-resolution,'' \emph{{IEEE} Trans. Multim.}, vol.~24, pp. 3506--3519, 2022.

\bibitem{NYU}
N.~Silberman, D.~Hoiem, P.~Kohli, and R.~Fergus, ``Indoor segmentation and support inference from {RGBD} images,'' in \emph{Proc. Eur. Conf. Comput. Vis. (ECCV)}, 2012, pp. 746--760.

\bibitem{FDSR}
L.~He, H.~Zhu, F.~Li, H.~Bai, R.~Cong, C.~Zhang, C.~Lin, M.~Liu, and Y.~Zhao, ``Towards fast and accurate real-world depth super-resolution: Benchmark dataset and baseline,'' in \emph{Proc. IEEE Conf. Comput. Vis. Pattern Recognit. (CVPR)}, 2021, pp. 9229--9238.

\bibitem{Lu}
S.~Lu, X.~Ren, and F.~Liu, ``Depth enhancement via low-rank matrix completion,'' in \emph{Proc. IEEE Conf. Comput. Vis. Pattern Recognit. (CVPR)}, 2014, pp. 3390--3397.

\bibitem{middle2001}
D.~Scharstein and R.~Szeliski, ``A taxonomy and evaluation of dense two-frame stereo correspondence algorithms,'' \emph{Int. J. Comput. Vis.}, vol.~47, no. 1-3, pp. 7--42, 2002.

\bibitem{middle2006}
H.~Hirschm{\"{u}}ller and D.~Scharstein, ``Evaluation of cost functions for stereo matching,'' in \emph{Proc. IEEE Conf. Comput. Vis. Pattern Recognit. (CVPR)}, 2007, pp. 1--8.

\bibitem{middle2014}
D.~Scharstein, H.~Hirschm{\"{u}}ller, Y.~Kitajima, G.~Krathwohl, N.~Nesic, X.~Wang, and P.~Westling, ``High-resolution stereo datasets with subpixel-accurate ground truth,'' in \emph{Proc. 36th German Conf. Pattern Recognit.}, 2014, pp. 31--42.

\bibitem{middle2005}
D.~Scharstein and C.~Pal, ``Learning conditional random fields for stereo,'' in \emph{Proc. IEEE Conf. Comput. Vis. Pattern Recognit. (CVPR)}, 2007, pp. 1--8.

\bibitem{DJFR}
Y.~Li, J.~Huang, N.~Ahuja, and M.~Yang, ``Joint image filtering with deep convolutional networks,'' \emph{{IEEE} Trans. Pattern Anal. Mach. Intell.}, vol.~41, no.~8, pp. 1909--1923, 2019.

\bibitem{SDF}
B.~Ham, M.~Cho, and J.~Ponce, ``Robust guided image filtering using nonconvex potentials,'' \emph{{IEEE} Trans. Pattern Anal. Mach. Intell.}, vol.~40, no.~1, pp. 192--207, 2018.

\bibitem{SVLMR}
J.~Pan, J.~Dong, J.~S.~J. Ren, L.~Lin, J.~Tang, and M.~Yang, ``Spatially variant linear representation models for joint filtering,'' in \emph{Proc. IEEE Conf. Comput. Vis. Pattern Recognit. (CVPR)}, 2019, pp. 1702--1711.

\bibitem{acm2021}
J.~Tang, X.~Chen, and G.~Zeng, ``Joint implicit image function for guided depth super-resolution,'' in \emph{Proc. ACM Int. Conf. Multim. (ACM MM)}, 2021, pp. 4390--4399.

\bibitem{22cvpr}
Z.~Zhao, J.~Zhang, S.~Xu, Z.~Lin, and H.~Pfister, ``Discrete cosine transform network for guided depth map super-resolution,'' in \emph{Proc. IEEE Conf. Comput. Vis. Pattern Recognit. (CVPR)}, 2022, pp. 5697--5707.

\bibitem{crm/JEI16/underwater}
C.~Li, J.~Guo, B.~Wang, R.~Cong, Y.~Zhang, and J.~Wang, ``Single underwater image enhancement based on color cast removal and visibility restoration,'' \emph{J. Electronic Imaging}, vol.~25, no.~3, p. 033012, 2016.

\bibitem{crm/spl21/underwater}
J.~Hu, Q.~Jiang, R.~Cong, W.~Gao, and F.~Shao, ``Two-branch deep neural network for underwater image enhancement in {HSV} color space,'' \emph{{IEEE} Signal Process. Lett.}, vol.~28, pp. 2152--2156, 2021.

\bibitem{RDB}
Y.~Zhang, Y.~Tian, Y.~Kong, B.~Zhong, and Y.~Fu, ``Residual dense network for image super-resolution,'' in \emph{Proc. IEEE Conf. Comput. Vis. Pattern Recognit. (CVPR)}, 2018, pp. 2472--2481.

\end{thebibliography}
}
\end{document}